\pdfoutput=1

\documentclass[11pt]{article}

\usepackage[final]{acl}

\usepackage{times}
\usepackage{latexsym}

\usepackage[T1]{fontenc}

\usepackage[utf8]{inputenc}

\usepackage{microtype}

\usepackage{inconsolata}

\usepackage{graphicx}
\usepackage{amsmath}
\usepackage{booktabs}
\usepackage{tabularx}
\usepackage{float}  
\usepackage{adjustbox}
\usepackage{longtable}
\title{Linguistic and Embedding-Based Profiling of Texts Generated by Humans and Large Language Models}

\author{Sergio E. Zanotto\textsuperscript{1} \and
  Segun Aroyehun\textsuperscript{2}\\
  \textsuperscript{1}Department of Linguistics \& Cluster of Excellence ``The Politics of Inequality''\\ 
  University of Konstanz \\
  \texttt{sergio.zanotto@uni-konstanz.de}\\
   \textsuperscript{2}Department of Politics and Public Administration\\ 
   University of Konstanz\\
   \texttt{segun.aroyehun@uni-konstanz.de}
  }

\begin{document}
\maketitle
\begin{abstract}
The rapid advancements in large language models (LLMs) have significantly improved their ability to generate natural language, making texts generated by LLMs increasingly indistinguishable from human-written texts. While recent research has primarily focused on using LLMs to classify text as either human-written or machine-generated texts, our study focuses on characterizing these texts using a set of linguistic features across different linguistic levels such as morphology, syntax, and semantics. We select a dataset of human-written and machine-generated texts spanning 8 domains and produced by 11 different LLMs. We calculate different linguistic features such as dependency length and emotionality, and we use them for characterizing human-written and machine-generated texts along with different sampling strategies, repetition controls, and model release dates. Our statistical analysis reveals that human-written texts tend to exhibit simpler syntactic structures and more diverse semantic content. Furthermore, we calculate the variability of our set of features across models and domains. Both human- and machine-generated texts show stylistic diversity across domains, with human-written texts displaying greater variation in our features. Finally, we apply style embeddings to further test variability among human-written and machine-generated texts. Notably, newer models output text that is similarly variable, pointing to a homogenization of machine-generated texts.

\end{abstract}

\section{Introduction}

The rapid advancements in language models have significantly improved their ability to generate natural language, making machine-generated texts (MGT) increasingly indistinguishable from human-written texts (HWT). Indeed, recent studies indicate that disinformation produced by state-of-the-art large language models (LLMs) is often perceived as more credible than that created by humans \citep{spitale2023ai}. This evolution has highlighted the importance of identifying MGT due to legitimate concerns about the potential for malicious actors to disseminate false information, as well as the broader need to uphold trust and authenticity across online platforms \citep{chakravarthi2025proceedings,li-etal-2024-team, sarvazyan2023overview}. Best systems for this task, often referred to as Human/Machine Authorship Attribution, all imply the use of LLMs, and different studies show marginal gains of leveraging stylometric features combined with LLMs for successfully achieving this task \citep{alecakir-etal-2024-groningen, wang-etal-2024-semeval-2024}.  

In this study, we utilize the RAID dataset \citep{dugan-etal-2024-raid}, a large-scale corpus designed for testing detection tools for machine-generated text and human-written text. This work focuses on analyzing different levels of linguistic analysis, such as morphology, syntax, and semantics, in order to characterize MGT from 11 different models with 4 different decoding strategies, along with the release date of the models. Thus, we analyze a set of representative linguistic features to distinguish human‐written texts (HWT) from machine‐generated texts (MGT), potentially uncovering distinctive linguistic patterns in a multi-domain setting (across 8 different domains). Moreover, we focus on linguistic variability over time to check whether there are patterns of linguistic homogenization in newer LLMs \citep{sourati2025shrinking, padmakumar2023does}. 

We test the following set of linguistic features per level of linguistic analysis: Textual level (Text Length, Sentence Length), Morphology (Morphological Complexity Index for Verbs and Nouns), Syntax (Dependency Tree Depth, Dependency Length), Lexical Level (Word Prevalence, Type-Token Ratio), Semantics (Semantic Similarity), Emotionality. We consider these features as representative of different levels of linguistic analysis of our interest, ensuring overlap with previous work on distinguishing between HWT and MGT \citep{zanotto2024human, guo2023close, uchendu-etal-2020-authorship}. We do not aim to test all possible linguistic features tested in other research on Human/Machine authorship attribution tasks \citep[e.g.,][]{simon2023using, uchendu-etal-2020-authorship}. 

We provide statistical results for Human/Machine authorship attribution for each of these features. Moreover, we train a binary logistic classifier using these features to distinguish between HWT and MGT. We use the logistic classifier to analyze feature importance to get an overall picture of the impact of these features in characterizing HWT and MGT.

To assess variability in linguistic representations across models and domains, we calculate the standard deviation of the Euclidean distance for our extracted features. Additionally, we map language models to their release dates to analyze temporal trends in model development. This framework highlights differences between models and reveals the impact of decoding strategies and domain linguistic constraints.

Furthermore, we apply a style embedding model \citep{patel2025styledistancestrongercontentindependentstyle} that has been used for authorship attribution to represent text styles, enabling further comparison of variability in HWT and MGT.

Our main contributions are: (i) We analyze linguistic differences between human-written texts (HWT) and machine-generated texts (MGT) with linguistic features across different linguistic levels such as morphology, syntax, and semantics to examine variability across different models and domains. We show how recent models tend to have similar linguistic variability, pointing to a risk of homogenization of texts. (ii) We employ a logistic classifier to identify different linguistic patterns between HWT and MGT by performing a feature importance analysis. (iii) We further use style embeddings to compare HWT and MGT showing that recent models tend to exhibit similar style variation within themselves, underlying how chat models output texts with more similar characteristics to HWT than their non-chat models.

\section{Related Work}

Scholars have explored various approaches to tackle the challenge of distinguishing between human-written and machine-generated texts. This task, often referred to as Human/Machine Authorship Attribution \citep{alecakir-etal-2024-groningen}, involves detecting whether a text is produced by a human or a generative language model, or attributing authorship among different models.

The Human/Machine attribution of authorship to a text carries significant social relevance, especially in areas such as fake news detection \citep{kumarage2023stylometric, jawahar-etal-2020-automatic}. The need for explainability becomes particularly important when engaging with a broad audience of non-experts, who may not have the means to access or comprehend detection models \citep{gehrmann-etal-2019-gltr}. As a result, numerous studies have focused on identifying human-explainable features that can differentiate between machine-generated (MGT) and human-written texts (HWT) \citep[e.g.,][]{dugan2023real, guo2023close, kumarage2023stylometric, uchendu-etal-2020-authorship}. To achieve this, researchers have employed diverse analytical approaches, including stylometric analysis \citep{kumarage2023stylometric, ma2023ai}, qualitative assessments \citep{guo2023close,gehrmann-etal-2019-gltr}, and linguistic feature analysis \citep{wang-etal-2024-m4, uchendu-etal-2020-authorship, ferracane-etal-2017-leveraging}, to diverse corpora, contexts, and generation tasks. Notably, different studies highlight the difficulties encountered by humans in distinguishing machine-generated texts from human-written texts, but little has been done to address the real-life consequences of this issue. \citep[e.g.,][]{chakraborty2023possibilities, jakesch2023human, fraser2024detecting}.

Moreover, classical machine learning algorithms, such as logistic regression, have been employed to train models on bag-of-words features to differentiate between HWT and MGT \citep{solaiman2019release, ippolito-etal-2020-automatic}. Other traditional methods leverage linguistic features, including POS-tags \citep{ferracane-etal-2017-leveraging}, surface-level features such as readability indexes or punctuation marks \citep{doughman2024exploring, malviya2025skdu}, topic modeling \citep{seroussi2014authorship}, sentiment analysis \citep{hossen2025unmasking}, and LIWC (Linguistic Inquiry and Word Count) features to provide deeper insights into the characteristics of MGT \citep{uchendu-etal-2020-authorship, li-etal-2014-towards}. HWT tend to be longer, express more sentiment polarity, especially negative sentiment, and show greater variability in discourse structures compared to MGT \citep[e.g.,][]{kim-etal-2024-threads, zanotto2024human}.

With the advent of Large Language Models (LLMs), fine-tuned models such as RoBERTa have achieved state-of-the-art performance in many tasks \citep{crothers2023machine, jawahar-etal-2020-automatic}. Thus, different shared tasks such as SemEval and IberLEF focus on the creation of the best models for tackling Human/Machine Authorship Attribution, with LLM-based systems always reaching the top positions \citep{wang-etal-2024-semeval-2024, sarvazyan2023overview}. Indeed, classifiers solely based on stylometric features reach poor accuracy in distinguishing Human/Machine authorship \citep{alecakir-etal-2024-groningen, sharma-mansuri-2024-team}, with marginal gains when combined with LLMs \citep{sharma-mansuri-2024-team}. 
Models leveraging Transformers encoders with token-level probabilistic features offer state-of-the-art detection capabilities \citep{sarvazyan-etal-2024-genaios, mitchell2023detectgpt}. 
Despite providing literature on classifiers for Human/Machine authorship attribution, the focus of our study is to examine possible characteristics that distinguish HWT and MGT on different levels of linguistic analysis. Few studies show that MGT tend to exhibit shorter texts on average, with lower emotional content and syntactically more complex structures compared to HWT \citep{zanotto2024human, guo2023close}. However, they overlook distinctions across models, decoding strategies, or stylistic variability. In doing this, we test variability among models and provide results that possibly point to the tendencies of recent models to exhibit less variability in their output, possibly due to the recent practice of training on machine-generated data \citep{shumailov2023curse}.

\section{Data}

For our research, we use the RAID dataset \citep{dugan-etal-2024-raid}, a large-scale corpus comprising over 6.2 million text generations, developed to evaluate detection methods for outputs from language models. The training set of the dataset is constructed from 13,371 human-written documents sampled across eight diverse domains --- Abstracts, Books, News, Poetry, Recipes, Reddit, Reviews, and Wikipedia --- offering a broad spectrum of human-written and machine-generated English texts for linguistic analysis. For each document, a corresponding generation prompt is dynamically created using either ``Chat'' or ``Non-Chat'' templates, ensuring compatibility with the target language model. All the models use the same specific set of prompts (refer to Section \ref{PE} in Appendix for examples). ``Chat'' models are fine‐tuned (often via supervised fine‐tuning and reinforcement learning) on multi‐turn interactions with humans. ``Non-Chat'' models are trained for next‐token prediction on unstructured text. Texts are generated using eleven models, namely GPT-2 XL, GPT-3, GPT-4, ChatGPT, Mistral-7B, Mistral-7B (Chat), MPT-30B, MPT-30B (Chat), LLaMA 2 70B (Chat), Cohere, and Cohere (Chat), under four decoding strategies (i.e., greedy decoding and sampling decoding paired with the presence, where possible, or absence of repetition penalties). Greedy models are more deterministic and repetitive, while sampling models introduce randomness to make the response more diverse, whereas the presence of a penalty on repetition avoids excessive looping and redundancy. Note that we exclude texts generated using adversarial attacks from the original dataset. Hence, the selected subset of the dataset is designed to reflect realistic generation scenarios, providing a robust source for analyzing linguistic similarities and differences between machine-generated and human-written texts. In the end, we consider a total amount of 467,985 texts. 

\section{Methodology}
We extract a range of linguistic features to analyze differences between human-written and machine-generated texts. These features include text length, sentence length,  morphological complexity index, dependency length, dependency depth, word prevalence, type-token ratio, semantic similarity, and emotionality. For each feature, we compute the mean and standard error of the mean for human-written and machine-generated texts and assess similarity between human and model distributions using the Mann-Whitney U-test \citep{mcknight2010mann}. To further explore stylistic variability, we apply a style embedding model to extract a representation of the writing style of each text, enabling a comparison of style consistency across human-written and machine-generated texts. We train a logistic classifier on all extracted linguistic features to predict text authorship based on linguistic features. Additionally, we assess variability across models and domains by computing the standard deviation of the Euclidean distance for all linguistic features across HWT and MGT. We employ Principal Component Analysis (PCA) for dimensionality reduction using style embeddings to quantify variability within models and domains. All codes are available at \url{https://github.com/Sergio-E-Zanotto/LingAIHUman}.

\subsection{Features Collection}
We present here in detail the linguistic features that we calculate for characterizing human-written (HWT) and machine-generated texts (MGT).

\textbf{Text Length}: \texttt{Text Length} is calculated by counting the total number of alphanumerical tokens in a text using the tokenizer in SpaCy ``en\_core\_web\_sm''. We average the length of all texts produced by humans or models with their own decoding and penalty settings. Text length provides shallow linguistic information on possible differences between humans and models and between different generation settings of models.

\textbf{Sentence Length}: \texttt{Sentence Length} is calculated by counting the total number of alphanumerical tokens per sentence, normalized by the total number of sentences produced by a specific model or human. This measure allows for a better fine-grained understanding of possible differences within text length, whereas two texts could have the same length, but a different number of sentences and therefore different sentence lengths.

\textbf{Morphological Complexity Index}: The \texttt{Morphological Complexity Index (MCI)} \citep{brezina2019morphological} measures the diversity of word forms associated with the same lemma, reflecting the morphological richness of a text. It is calculated by extracting lemmas and their word forms using spaCy ``en\_core\_web\_sm'', then randomly sampling subsets of five words to compute within-subset variety (how many unique word forms appear within a subset) and between-subset diversity (how different two subsets are). A higher MCI indicates greater morphological complexity, making it useful for analyzing linguistic richness and distinguishing between different registers and writing styles.

\textbf{Dependency Tree Depth}: \texttt{Dependency Tree Depth} is calculated using the dependency parser in SpaCy ``en\_core\_web\_sm''. It represents hierarchical syntactic complexity. We calculate the maximum depth of the dependency tree for each sentence, from the syntactic head to the lowest leaf node. A deeper tree suggests more complex sentence constructions (e.g., multiple layers of subordinate clauses), whereas a shallower tree suggests a simpler structure.

\textbf{Dependency Length}: \texttt{Dependency Length} is calculated using the dependency parser in SpaCy ``en\_core\_web\_sm''. It represents the distance in number of tokens between a word (dependent) and its syntactic head. We considered for this measure only the intervening words between the head and the last dependent word. This measure focuses on the linear arrangement of words. Longer dependency lengths can indicate that related words are spread farther apart, which reflects more complex syntactic structures.

\textbf{Word Prevalence}: \texttt{Word prevalence} refers to the proportion of people who recognize and understand a given word. It is a measure of lexical familiarity, capturing how widely known a word is within a population. To calculate word prevalence, we used the English Word Prevalence dataset \citep{brysbaert2019word}, which provides prevalence scores based on a large-scale crowdsourcing study involving over 220,000 participants. Each word's prevalence score represents the percentage of respondents who reported knowing the word. We computed the average prevalence per text by tokenizing the text using SpaCy ``en\_core\_web\_sm'', extracting word forms and their lemmas, and matching them to their prevalence scores from the dataset. The prevalence scores were then averaged across all words in the text. This measure is useful for analyzing word difficulty, lexical accessibility, and text comprehensibility. 

\textbf{Type-Token Ratio}: \texttt{Type\_token\_ratio} is calculated by counting the number of unique tokens and dividing it by the total number of tokens per text. We include this feature to verify if HWT and MGT tend to have similar lexical diversity in their texts. 

\textbf{Semantic Similarity}: \texttt{Semantic Distance} is calculated using sentence embeddings derived from the Sentence Transformer model ``paraphrase-MiniLM-L6-v2'' \citep{reimers-2019-sentence-bert}. We calculate the cosine similarity in pair-wise sentence comparisons and averaged the distance for each document. This feature describes the semantic content of a text in terms of consistency, as in \citet{beaty2021automating}. 

\textbf{Emotionality}: \texttt{Emotionality} is calculated using the NRC Emotion Intensity Lexicon \citep{Mohammad13}. The lexicon includes 8 different emotions: anger, disgust, fear, sadness, joy, anticipation, surprise, trust. We consider anger, fear and sadness as negative emotions, while we take joy as the only representative positive emotion in the lexicon following \citet{aroyehun2023leia}. We compute emotional load as the proportion of positive and negative emotion words in a given text. This feature helps to describe potential differences in emotional content between HWT and MGT, as in \citet{guo2023close}.

\textbf{Style Embeddings}: We compute style embeddings using the StyleDistance model \citep{patel2025styledistancestrongercontentindependentstyle}. This model generates style embeddings with a dimension of 768 such that texts with similar stylistic features are closer in embedding space, independent of content. We use these embeddings to assess whether models and humans exhibit different stylistic variability in their generated texts. We selected StyleDistance \citep{patel2025styledistancestrongercontentindependentstyle} because it is a state-of-the-art model explicitly designed to generate content-independent style embeddings. The model has been shown to generalize well to real-world benchmarks and outperform leading style representation methods across multiple downstream tasks. Evaluations on style identification, style transfer, and authorship verification demonstrate that StyleDistance captures stylistic properties robustly, including under out-of-domain and out-of-distribution conditions.
We use the resulting 768-dimensional embeddings as a latent representation of linguistic style, providing an additional analysis alongside our main feature-based approach. While we do not assume a one-to-one correspondence between embedding dimensions and linguistic features, these embeddings allow us to assess whether global stylistic patterns across models and domains align with those revealed through explicit linguistic features.

\subsection{Models and Domain Variability}
For each model and domain, we first compute the centroid by taking the mean of the values of our linguistic features. Then, for each individual observation in that model or domain, we calculate the Euclidean distance from its feature vector to the centroid. The overall variability for the models and domains is quantified as the standard deviation of these distances. These standard deviations reflect how spread out the observations are in the feature space relative to the centroid. 
Moreover, we apply dimensionality reduction using PCA to the style embeddings to match the linguistic feature count (10 dimensions). We compute variability within domains and models for the style embeddings by calculating their centroids, Euclidean distances, and standard deviations. Finally, we map models to their release dates to identify a possible trend in the evolution of models with respect to our set of linguistic features and style embeddings.

\subsection{Feature Importance via a Logistic Classifier}

We train a binary logistic regression classifier with a set of linguistic features to characterize via feature importance analysis human-written texts (label: 1) from machine-generated texts (label: 0). To ensure a balanced dataset, we downsample machine-generated texts, resulting in a training set of 21,376 documents and a test set of 5,344. Rather than comparing with existing methods, our focus is on assessing feature importance to get an overall picture of the impact of these features in characterizing HWT and MGT. We do not aim to build a state‑of‑the‑art detection classifier nor to compare classification results with existing literature.

\section{Results} \label{results}
\begin{figure*}[ht]
    \centering
    \includegraphics[width=\textwidth]{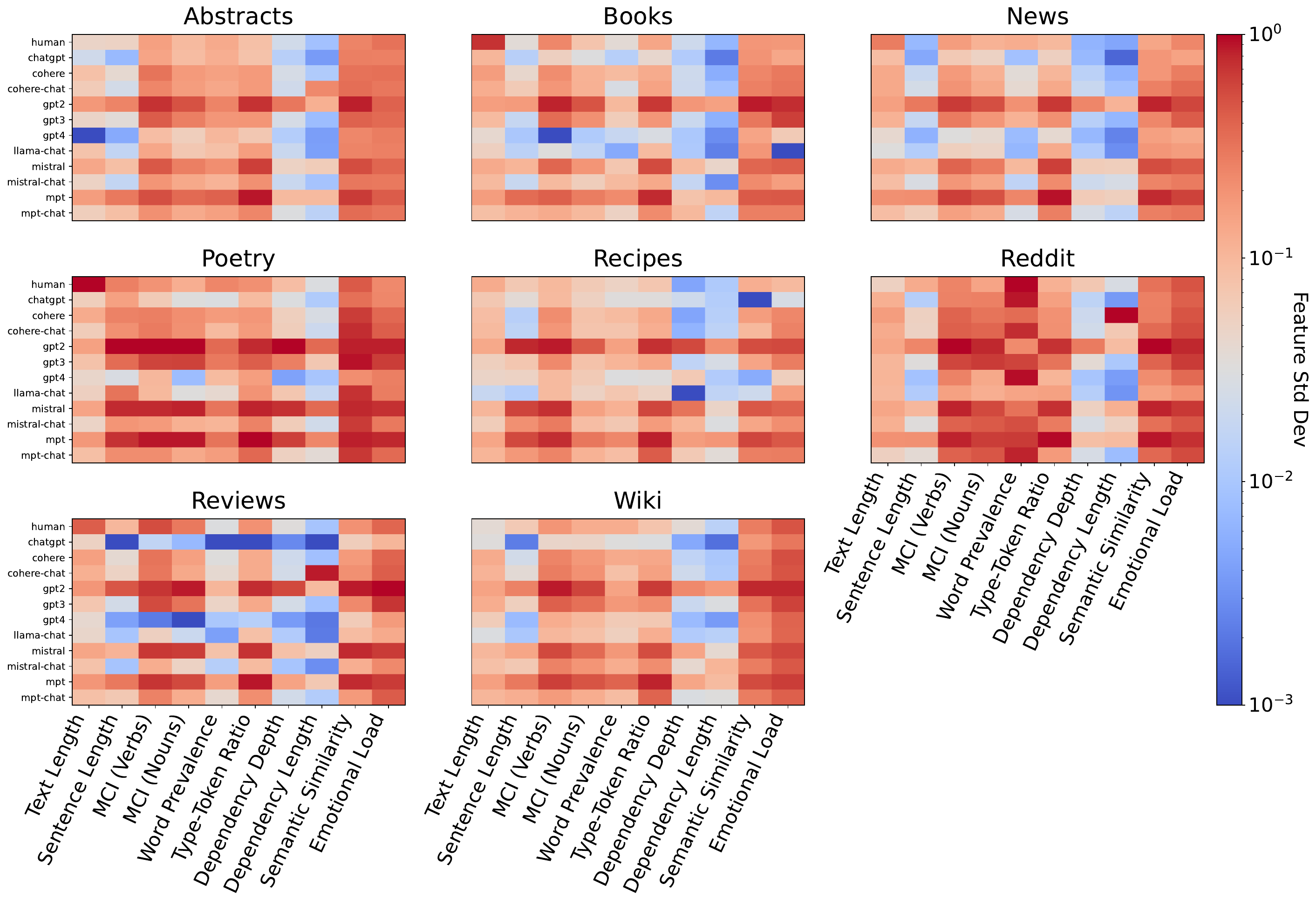}
    \captionsetup{width=\textwidth}
    \caption{Multi-panel heatmap displaying log-normalized feature standard deviation across different text domains and models}
    \label{fig:multi_panel_heatmap_std}
\end{figure*}

In this section, we present the linguistic features used to analyze human-written texts (HWT) and machine-generated texts (MGT). These features capture differences in text structure, morphological complexity,  syntactic complexity, lexical diversity, and emotionality. We show the results and feature importance from the classifier based on the calculated linguistic features. Figure \ref{fig:multi_panel_heatmap_std} illustrates the variation in linguistic features across domains, showing that MGT exhibit comparatively low variability. Section \ref{LFV} in the Appendix presents further visualizations and discussions of observed patterns across domains. These plots highlight the influence of genre-specific constraints on different domains. In more creative domains such as poetry and books, human-written texts tend to be longer and semantically more diverse. Across all domains, human-written texts consistently show simpler syntactic structures, while maintaining a relatively stable use of common words, as measured by Word Prevalence. Notably, this measure varies among HWT between Reddit and news: human-written texts on Reddit show the highest use of common words, whereas News articles show the lowest.

\begin{table*}[t]
    \centering
    \resizebox{\textwidth}{!}{%
    \begin{tabular}{lrrrrrrrrrr}
    \toprule
                    model &   Text Length & Sentence Length &    MCI (VERBs) &    MCI (NOUNs) & Word Prevalence &    Type-Token Ratio & Dependency Depth & Dependency Length & Semantic Similarity & Emotionality \\
    \midrule
    \textbf{human} & 300.61 ± 2.766 &       23.53 ± 0.134 & 8.73 ± 0.011 & 8.73 ± 0.006 &       2.27 ± 0.001 & 0.64 ± 0.001 &    6.83 ± 0.017 &              2.41 ± 0.006 &        0.30 ± 0.001 &    0.38 ± 0.001 \\
        chatgpt\_greedy\_no & \textbf{260.35 ± 0.950} &       19.29 ± 0.065 & 8.60 ± 0.009 & 8.71 ± 0.006 &       2.28 ± 0.001 & 0.62 ± 0.001 &    \textbf{6.70 ± 0.012} &              2.27 ± 0.004 &        0.38 ± 0.001 &    0.37 ± 0.001 \\
      chatgpt\_sampling\_no & 272.73 ± 0.960 &       20.00 ± 0.077 & \textbf{8.70 ± 0.008} & 8.80 ± 0.006 &       2.28 ± 0.001 & 0.63 ± 0.001 &    \textbf{6.75 ± 0.011} &              \textbf{2.32 ± 0.004} &        0.37 ± 0.001 &    \textbf{0.38 ± 0.001} \\
    cohere-chat\_greedy\_no & 189.48 ± 0.743 &       21.47 ± 0.106 & 7.91 ± 0.012 & 8.29 ± 0.008 &       2.28 ± 0.001 & 0.63 ± 0.001 &    6.72 ± 0.013 &              \textbf{2.44 ± 0.038} &        0.33 ± 0.001 &    0.34 ± 0.001 \\
  cohere-chat\_sampling\_no & 190.51 ± 0.772 &       21.58 ± 0.105 & 7.92 ± 0.012 & 8.31 ± 0.008 &       2.28 ± 0.001 & \textbf{0.64 ± 0.001} &    6.73 ± 0.014 &              2.58 ± 0.104 &        0.33 ± 0.001 &    0.34 ± 0.001 \\
         cohere\_greedy\_no & 234.41 ± 0.911 &       21.29 ± 0.106 & 8.30 ± 0.012 & 8.50 ± 0.007 &       2.28 ± 0.000 & 0.61 ± 0.001 &    6.58 ± 0.013 &              2.80 ± 0.144 &        0.32 ± 0.001 &    0.36 ± 0.001 \\
       cohere\_sampling\_no & 239.54 ± 0.934 &       22.11 ± 0.132 & 8.33 ± 0.012 & 8.54 ± 0.007 &       2.27 ± 0.001 & 0.63 ± 0.001 &    6.64 ± 0.014 &              2.61 ± 0.076 &        0.31 ± 0.001 &    0.37 ± 0.001 \\
           gpt2\_greedy\_no & 364.78 ± 0.678 &       79.62 ± 0.939 & 2.98 ± 0.023 & 5.02 ± 0.022 &       2.26 ± 0.001 & 0.10 ± 0.001 &   16.59 ± 0.242 &              5.15 ± 0.107 &        0.59 ± 0.003 &    0.17 ± 0.001 \\
          gpt2\_greedy\_yes & 243.15 ± 1.059 &       25.84 ± 0.407 & 8.91 ± 0.018 & 8.38 ± 0.012 &       2.29 ± 0.000 & 0.67 ± 0.002 &    7.67 ± 0.090 &              2.78 ± 0.054 &        0.29 ± 0.001 &    0.35 ± 0.001 \\
         gpt2\_sampling\_no & 314.53 ± 0.904 &       \textbf{22.45 ± 0.107} & 8.98 ± 0.014 & 8.79 ± 0.008 &       2.29 ± 0.000 & 0.58 ± 0.001 &    6.76 ± 0.018 &              2.56 ± 0.006 &        0.29 ± 0.001 &    0.37 ± 0.001 \\
        gpt2\_sampling\_yes & 297.14 ± 1.062 &       26.90 ± 0.085 & 9.38 ± 0.015 & 8.93 ± 0.010 &       2.30 ± 0.000 & 0.78 ± 0.001 &    7.37 ± 0.014 &              2.55 ± 0.005 &        0.25 ± 0.001 &    0.39 ± 0.001 \\
           gpt3\_greedy\_no & 136.86 ± 0.659 &       19.95 ± 0.182 & 6.61 ± 0.017 & 7.34 ± 0.015 &       2.27 ± 0.001 & 0.63 ± 0.001 &    6.54 ± 0.035 &              2.38 ± 0.012 &        0.33 ± 0.001 &    0.29 ± 0.001 \\
         gpt3\_sampling\_no & 139.58 ± 0.672 &       19.42 ± 0.112 & 6.84 ± 0.017 & 7.50 ± 0.014 &       \textbf{2.27 ± 0.001 } & 0.66 ± 0.001 &    6.44 ± 0.023 &              2.35 ± 0.008 &        0.32 ± 0.001 &    0.30 ± 0.001 \\
           gpt4\_greedy\_no & 267.44 ± 0.782 &       18.21 ± 0.048 & 8.51 ± 0.009 & 8.81 ± 0.004 &       2.28 ± 0.001 & 0.60 ± 0.001 &    6.54 ± 0.015 &              2.37 ± 0.005 &        0.35 ± 0.001 &    0.37 ± 0.001 \\
         gpt4\_sampling\_no & 289.58 ± 0.816 &       19.14 ± 0.038 & 8.69 ± 0.008 & 8.94 ± 0.004 &       2.27 ± 0.001 & 0.65 ± 0.001 &    6.56 ± 0.011 &              2.44 ± 0.004 &        0.35 ± 0.001 &    0.39 ± 0.001 \\
                    
     llama-chat\_greedy\_no & 281.95 ± 0.547 &       23.19 ± 0.143 & 8.84 ± 0.007 & 8.85 ± 0.004 &       2.29 ± 0.000 & 0.57 ± 0.001 &    7.08 ± 0.014 &              2.48 ± 0.005 &        0.34 ± 0.001 &    0.38 ± 0.001 \\
    llama-chat\_greedy\_yes & 268.22 ± 0.572 &       23.62 ± 0.151 & 9.00 ± 0.007 & 8.93 ± 0.004 &       2.29 ± 0.000 & 0.66 ± 0.001 &    7.07 ± 0.014 &              2.41 ± 0.004 &        0.31 ± 0.001 &    0.38 ± 0.001 \\
   llama-chat\_sampling\_no & 281.03 ± 0.547 &       23.12 ± 0.140 & 8.83 ± 0.007 & 8.85 ± 0.004 &       2.28 ± 0.000 & 0.58 ± 0.001 &    7.05 ± 0.014 &              2.48 ± 0.005 &        0.34 ± 0.001 &    0.38 ± 0.001 \\
  llama-chat\_sampling\_yes & 265.09 ± 0.585 &       23.61 ± 0.138 & 8.97 ± 0.007 & 8.96 ± 0.004 &       2.29 ± 0.000 & 0.69 ± 0.001 &    7.09 ± 0.014 &              \textbf{2.40 ± 0.005} &        0.30 ± 0.001 &    0.39 ± 0.001 \\
   mistral-chat\_greedy\_no & 205.28 ± 0.634 &       23.17 ± 0.176 & 8.03 ± 0.012 & 8.41 ± 0.008 &       2.29 ± 0.001 & 0.59 ± 0.001 &    7.17 ± 0.028 &              \textbf{2.52 ± 0.021} &        0.38 ± 0.001 &    0.34 ± 0.001 \\
  mistral-chat\_greedy\_yes & 200.14 ± 0.608 &       21.79 ± 0.101 & 8.53 ± 0.009 & 8.65 ± 0.007 &       2.30 ± 0.001 & 0.72 ± 0.001 &    7.04 ± 0.013 &              2.36 ± 0.005 &        0.33 ± 0.001 &    0.36 ± 0.001 \\
 mistral-chat\_sampling\_no & 210.89 ± 0.663 &       22.07 ± 0.099 & 8.26 ± 0.010 & 8.55 ± 0.007 &       2.29 ± 0.001 & 0.62 ± 0.001 &    7.02 ± 0.014 &              2.49 ± 0.022 &        0.37 ± 0.001 &    0.36 ± 0.001 \\
mistral-chat\_sampling\_yes & 199.95 ± 0.698 &       23.51 ± 0.097 & 8.43 ± 0.011 & \textbf{8.67 ± 0.008} &       2.30 ± 0.001 & 0.77 ± 0.001 &    7.31 ± 0.014 &              2.39 ± 0.005 &        0.32 ± 0.001 &    0.37 ± 0.001 \\
        mistral\_greedy\_no & 316.24 ± 0.651 &       50.66 ± 0.718 & 5.20 ± 0.027 & 6.73 ± 0.021 &       2.26 ± 0.001 & 0.24 ± 0.001 &   10.94 ± 0.158 &              3.77 ± 0.088 &        0.44 ± 0.003 &    0.25 ± 0.001 \\
       mistral\_greedy\_yes & 208.68 ± 0.823 &       20.60 ± 0.150 & 8.86 ± 0.013 & 8.56 ± 0.008 &       2.29 ± 0.001 & 0.78 ± 0.001 &    6.60 ± 0.031 &              2.44 ± 0.030 &        0.28 ± 0.001 &    0.36 ± 0.001 \\
      mistral\_sampling\_no & 286.23 ± 0.743 &       21.29 ± 0.092 & 8.90 ± 0.011 & 8.81 ± 0.006 &       2.28 ± 0.001 & 0.60 ± 0.001 &    6.52 ± 0.012 &              2.54 ± 0.005 &        0.28 ± 0.001 &    0.37 ± 0.001 \\
     mistral\_sampling\_yes & 236.98 ± 0.869 &       31.47 ± 0.140 & 8.97 ± 0.011 & 8.80 ± 0.006 &       2.30 ± 0.000 & 0.85 ± 0.000 &    7.85 ± 0.018 &              2.68 ± 0.006 &        0.26 ± 0.001 &    \textbf{0.38 ± 0.001} \\
       mpt-chat\_greedy\_no & 157.84 ± 0.623 &       22.06 ± 0.113 & 7.58 ± 0.011 & 8.12 ± 0.009 &       2.29 ± 0.001 & 0.66 ± 0.001 &    7.09 ± 0.021 &              2.43 ± 0.007 &        0.38 ± 0.001 &    0.33 ± 0.001 \\
      mpt-chat\_greedy\_yes & 165.66 ± 0.766 &       26.20 ± 0.125 & 8.01 ± 0.012 & 8.46 ± 0.009 &       2.29 ± 0.001 & 0.87 ± 0.001 &    7.44 ± 0.019 &              2.47 ± 0.011 &        0.33 ± 0.001 &    0.36 ± 0.001 \\
     mpt-chat\_sampling\_no & 161.26 ± 0.648 &       22.42 ± 0.112 & 7.64 ± 0.011 & 8.18 ± 0.009 &       2.29 ± 0.001 & 0.67 ± 0.001 &    7.08 ± 0.018 &              2.42 ± 0.005 &        0.37 ± 0.001 &    0.34 ± 0.001 \\
    mpt-chat\_sampling\_yes & 180.93 ± 0.911 &       31.72 ± 0.194 & 8.01 ± 0.013 & 8.47 ± 0.010 &       2.28 ± 0.001 & 0.92 ± 0.001 &    7.66 ± 0.018 &              2.75 ± 0.013 &        0.31 ± 0.001 &    0.37 ± 0.001 \\
            mpt\_greedy\_no & 361.78 ± 0.569 &       43.87 ± 0.660 & 5.27 ± 0.026 & 6.73 ± 0.020 &       2.26 ± 0.001 & 0.20 ± 0.001 &    9.46 ± 0.133 &              3.52 ± 0.068 &        0.50 ± 0.002 &    0.25 ± 0.001 \\
           mpt\_greedy\_yes & 192.30 ± 0.945 &       48.82 ± 0.314 & 8.33 ± 0.016 & 8.33 ± 0.013 &       2.29 ± 0.001 & 1.00 ± 0.000 &    8.89 ± 0.025 &              3.55 ± 0.042 &        0.25 ± 0.001 &    0.36 ± 0.001 \\
          mpt\_sampling\_no & 349.96 ± 0.613 &       21.02 ± 0.088 & 9.23 ± 0.011 & 8.98 ± 0.006 &       2.28 ± 0.001 & 0.57 ± 0.001 &    6.53 ± 0.012 &              2.48 ± 0.008 &        0.27 ± 0.001 &    \textbf{0.38 ± 0.001} \\
         mpt\_sampling\_yes & 245.68 ± 1.194 &       47.13 ± 0.388 & 8.49 ± 0.021 & 8.36 ± 0.016 &       2.28 ± 0.001 & 1.00 ± 0.000 &    7.79 ± 0.025 &              3.53 ± 0.025 &        0.22 ± 0.001 &    0.36 ± 0.001 \\
    \bottomrule
    \end{tabular}%
    }
    \caption{Average values $\pm$ standard errors (SE) of calculated linguistic features per model in the dataset. The pattern of each model name is: model + decoding strategy + yes/no penalty on repetitions.
    Models corresponding to \textit{non-bold} entries are statistically different from human writing in the respective feature. \textit{Bold} entries are \textbf{not} statistically different based on the Mann-Whitney U-test ($p \geq 0.05$), indicating that the feature distribution in the machine-generated texts is not distinguishable from human-written texts. 
    }
    \label{tab:feature_statistics_news}
\end{table*}

\begin{figure}[t]
    \centering
    \includegraphics[width=\linewidth]{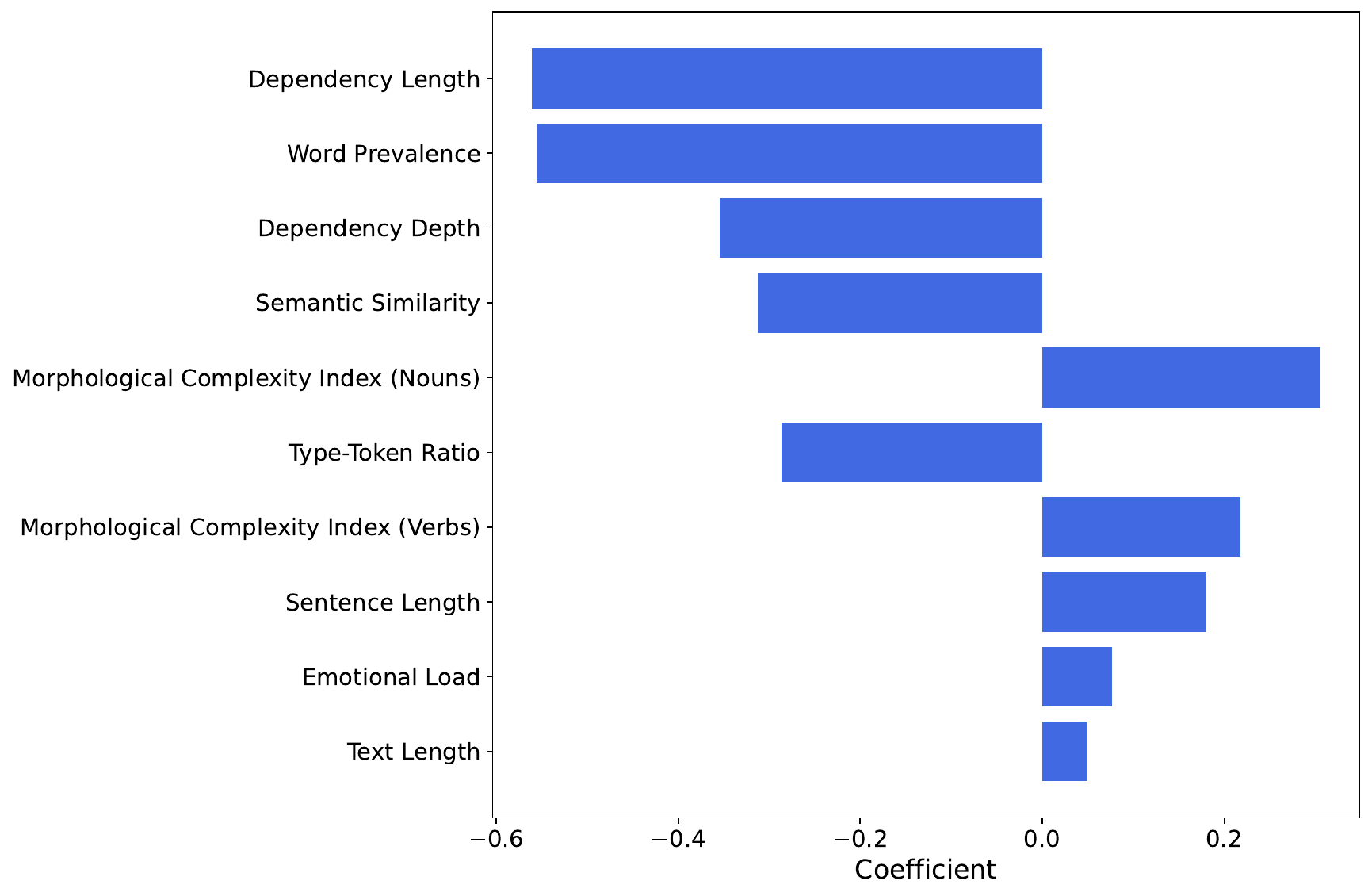}
    \caption{Feature importance for the logistic classifier across all domains. A positive coefficient means that as a feature value increases, the model is more likely to predict human authorship, while a negative coefficient indicates that higher values make the model less likely to predict human authorship.}
    \label{fig:feature_importance}
\end{figure}

Table \ref{tab:feature_statistics_news} shows the statistical analysis of our selected set of features for HWT and MGT in the entire dataset. Bold entries are those models that are statistically similar to humans for that feature. Different models and decoding strategies show different behaviours per feature in comparison to humans. We notice a tendency for chat models and for sampling strategies to output texts that are more similar to humans. Thus, we rely on the feature importance of a logistic classifier to extract further tendencies. Figure \ref{fig:feature_importance} illustrates the feature importance for the overall dataset. The classifier primarily relies on simpler syntactic structures (e.g., shorter dependency length and depth) and semantic properties (e.g., lower word prevalence, reduced semantic similarity, and a lower type-token ratio) for distinguishing between human-written and machine-generated texts. We present an analysis of each feature in the next section.

\subsection{How do HWT and MGT differ in terms of linguistic features?}

\textbf{Text Length}: Table \ref{tab:feature_statistics_news} shows the average length of HWT and MGT in the entire dataset. Previous studies found HWT to exhibit, on average, longer texts than MGT \citep{guo2023close}. In our study, this tendency is confirmed with a lower margin, since some models with the no repetition penalty setting generate longer texts on average than humans (e.g., gpt2). However, Section \ref{LFV} in the Appendix presents the feature analysis across domains. Notably, in the Poetry and Book domains, HWT are longer and show greater variability than MGT, likely reflecting the effect of more ``creative'' domains with less rigid genre-specific constraints. This may enable human authors to express individual creativity more distinctly than models.

\textbf{Sentence Length}: Table \ref{tab:feature_statistics_news} shows the average sentence length of HWT and MGT in the entire dataset. We do not notice a clear pattern that distinguishes humans from different models with decoding strategies or penalty control. Figure \ref{fig:feature_importance} shows that the classifier uses sentence length as a distinguishing feature, with a tendency for HWT to be longer than those of MGT.

\textbf{Morphological Complexity Index (MCI)}:
Table \ref{tab:feature_statistics_news} shows the MCI for nouns and the MCI for verbs across models in the dataset. Again, we notice that human-written texts tend to be somewhat in the middle between models with different decoding strategies and the presence of penalties (e.g., mpt). However, HWT rather tend to have a higher inflectional diversity for both verbs and nouns than MGT (See Figure \ref{fig:feature_importance}).

\textbf{Dependency Tree Depth}: 
Table \ref{tab:feature_statistics_news} shows the average syntactical dependency depth per model across the entire dataset. We notice that HWT tend to score lower in dependency depth than older models, while recent models such as GPT-4 show a more human-like amount of depth in the syntactic tree. This evolution is clear when compared to other studies in which the syntactic depth of HWT was lower than all the other models \citep{zanotto2024human}. 

\textbf{Dependency Length}: 
To further investigate the syntactic characteristics of human-written texts and machine-generated texts, Table \ref{tab:feature_statistics_news} shows the average dependency length per model across the entire dataset and complements the dependency depth results. Indeed, HWT score lower than older models like GPT-2 in dependency length and depth, while newer models produce similarly short or even shorter dependencies.  

\textbf{Word Prevalence}:
Table \ref{tab:feature_statistics_news} shows the average word prevalence per model across the dataset. While HWT score slightly lower than most models, the differences are minimal. Both HWT and MGT exceed a two z-score threshold, indicating that 98\% of the population is familiar with the words used.

\textbf{Type-Token Ratio}: 
Table \ref{tab:feature_statistics_news} depicts the average type-token ratio per model across the dataset. Human-written texts maintain a balanced ratio of around 0.6, indicating a balanced mix of unique words and repetitions. Models with a repetition penalty show higher type-token ratios than those without (e.g., mistral-chat).

\textbf{Semantic Similarity}: 
Table \ref{tab:feature_statistics_news} shows that human-written texts (HWT) exhibit a lower mean similarity than most models, indicating greater semantic diversity in human writing (See Figure \ref{fig:multi_panel_heatmap_mean} in Appendix). Repetition penalties help reduce redundancy, while sampling generates more diverse content than greedy decoding, as demonstrated with GPT-2.

\textbf{Emotionality}: 
In Table \ref{tab:feature_statistics_news}, we can notice how HWT score high in average emotionality in the dataset, but it does not score the highest, unlike what previous studies found \citep{guo2023close}. Indeed, differences between models seem to be driven by the decoding strategy and the presence of the penalty, where the majority of models with a sampling decoding strategy and a penalty on repetition score higher than the others (see Figure \ref{fig:multi_panel_heatmap_mean} in the Appendix). Looking at possible differences between emotional load for positive and negative emotions, Figure \ref{fig:raid_emotion_intensity} in the Appendix shows how there are some exceptions where HWT do not consistently score higher on negative emotions than MGT, contrary to previous findings \citep{guo2023close}.

\subsection{Domain Variability}

\begin{figure}[ht]
    \centering
    \includegraphics[width=\linewidth]{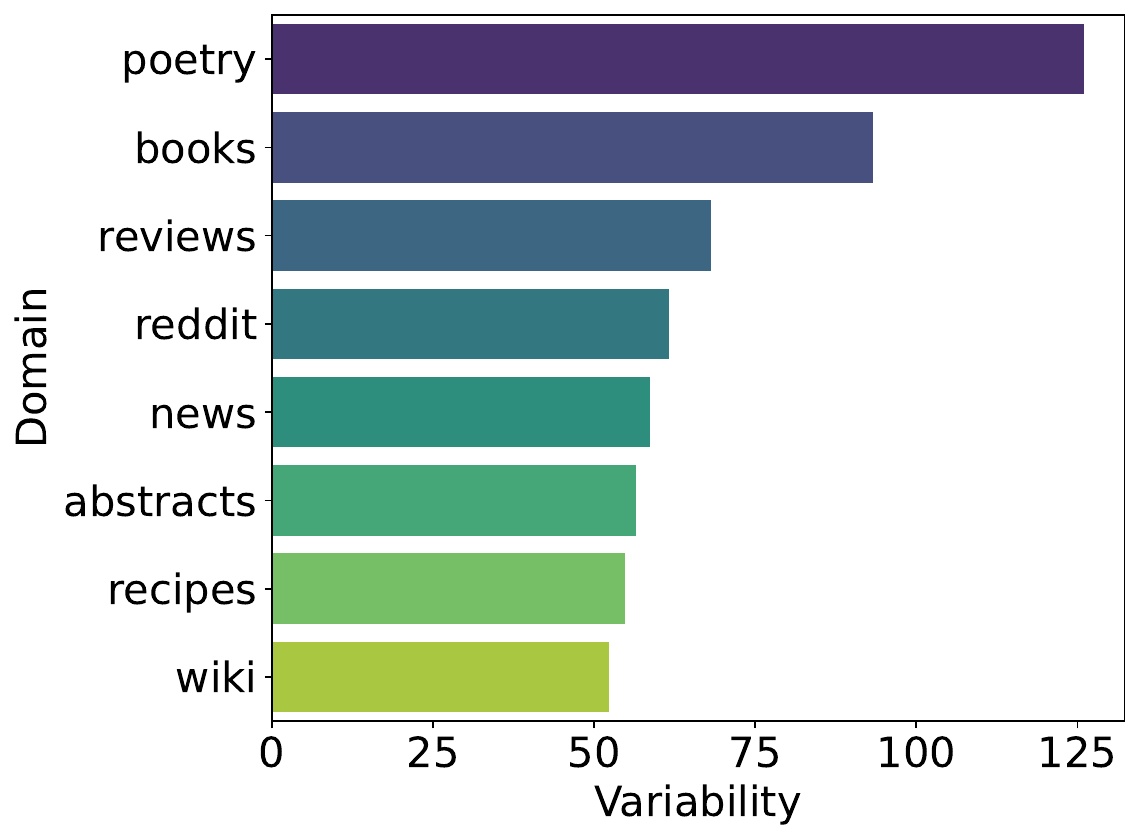}
    \caption{Domain variability with linguistic features in the entire dataset}
    \label{fig:domain_variability}
\end{figure}

Figure \ref{fig:domain_variability} shows the variability of linguistic features across domains. The results highlight variations in the linguistic feature space, reflecting the well-established influence of genre-specific constraints on linguistic variability \citep{biber2019register}. Indeed, more creative domains like poetry show more variability than less creative domains such as abstracts. Figure \ref{fig:style_domain_variability} in Appendix shows the variability based on style embeddings across domains. Style embeddings capture less variability in poetry than in other domains.
To better understand the linguistic features potentially driving these differences, an avenue for future work will be to extract the set of individual features used to train the style embeddings model and identify which ones are responsible for the different patterns. This alternative approach is beyond the scope of the current study.

\begin{figure}[ht]
    \centering
    \includegraphics[width=\linewidth]{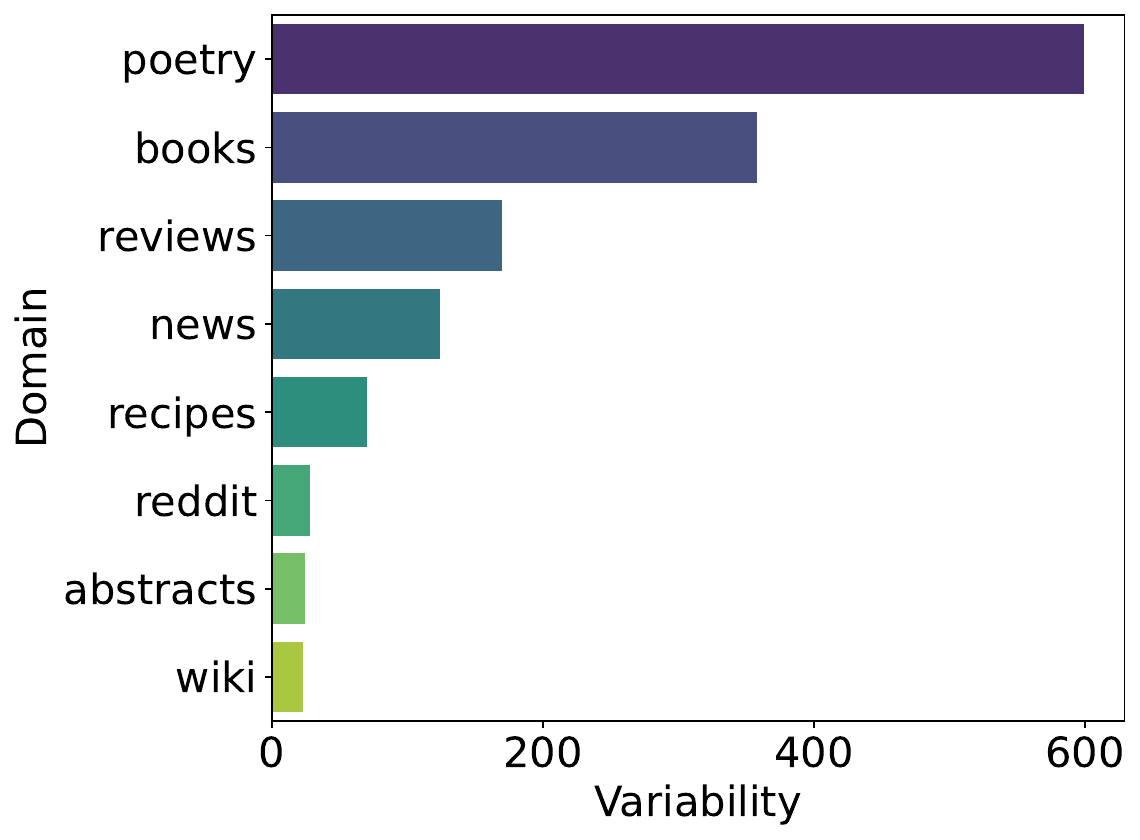}
    \caption{Domain variability with linguistic features in the entire dataset for humans}
    \label{fig:domain_variability_huamns}
\end{figure}

\begin{figure}[ht]
    \centering
    \includegraphics[width=\linewidth]{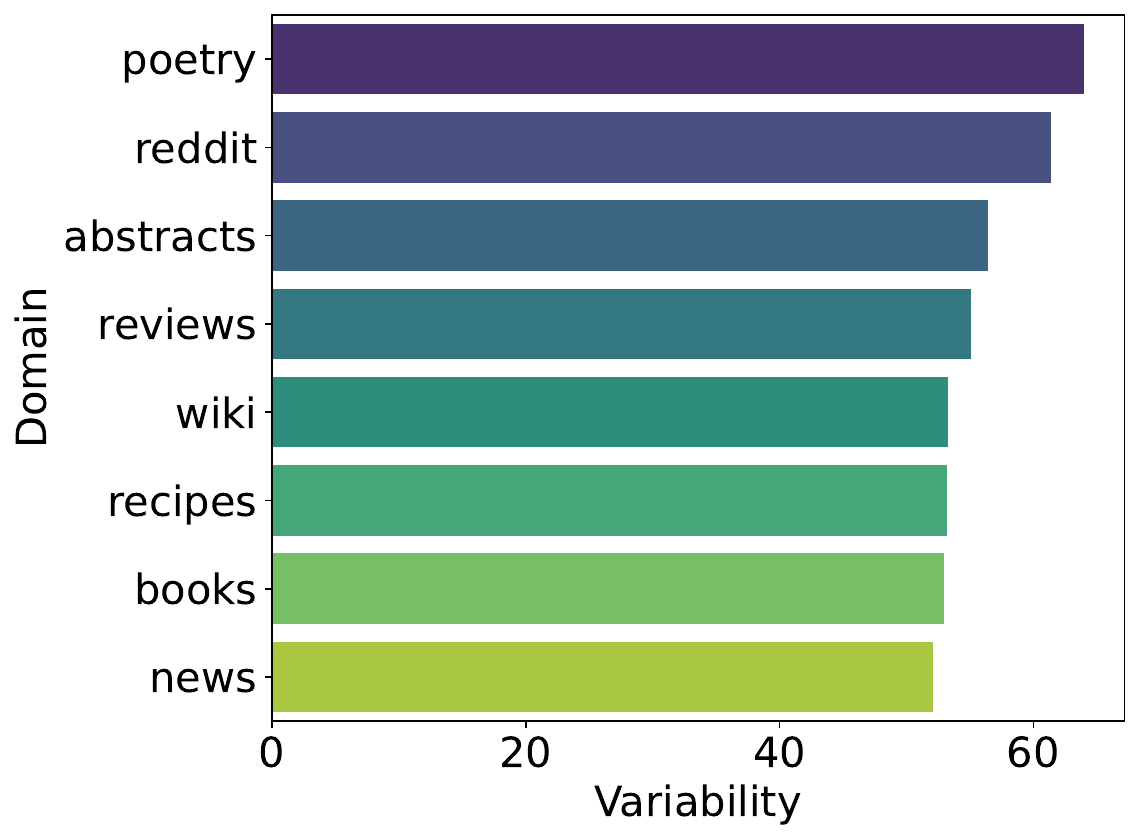}
    \caption{Domain variability with linguistic features in the entire dataset for non-humans}
    \label{fig:domain_variability_nonhumans}
\end{figure}

Moreover, we disentangle humans and models to understand if they behave differently in terms of variability for both our linguistic features and for the style embeddings. Figure \ref{fig:domain_variability_huamns} shows the domain variability of HWT in the linguistic feature space, while Figure \ref{fig:domain_variability_nonhumans} shows the domain variability of MGT in the linguistic feature space. HWT exhibit more variability than MGT across domains, arguably as they are more sensible to the well-known effect of different genre-specific constraints \citep{biber2019register}. Figure \ref{fig:domain_variability_huamns} shows how poetry is the most variable domain, arguably as the most ``creative'' domain of the dataset.
However, the domain variability of HWT and MGT appears more similar when using the style embeddings. Figure \ref{fig:style_domain_variability_huamns} in Appendix shows how Wiki is the most variable domain, in contrast to the linguistic features analysis. However, the variation in terms of standard deviation is comparable with the results based on the linguistic features. Figure \ref{fig:style_domain_variability_nonhumans} in Appendix shows poetry to be the least variable domain, and exhibits marginally lower variability across several domains, for example Reddit or Abstracts. We argue that these differences stem from the different features that are considered for the analyses: explicit linguistic features vs. latent style embedding representations. Further investigation would be needed to understand which dimensions can be attributed to the different observations.

\begin{figure}[ht]
    \centering
    \includegraphics[width=\linewidth]{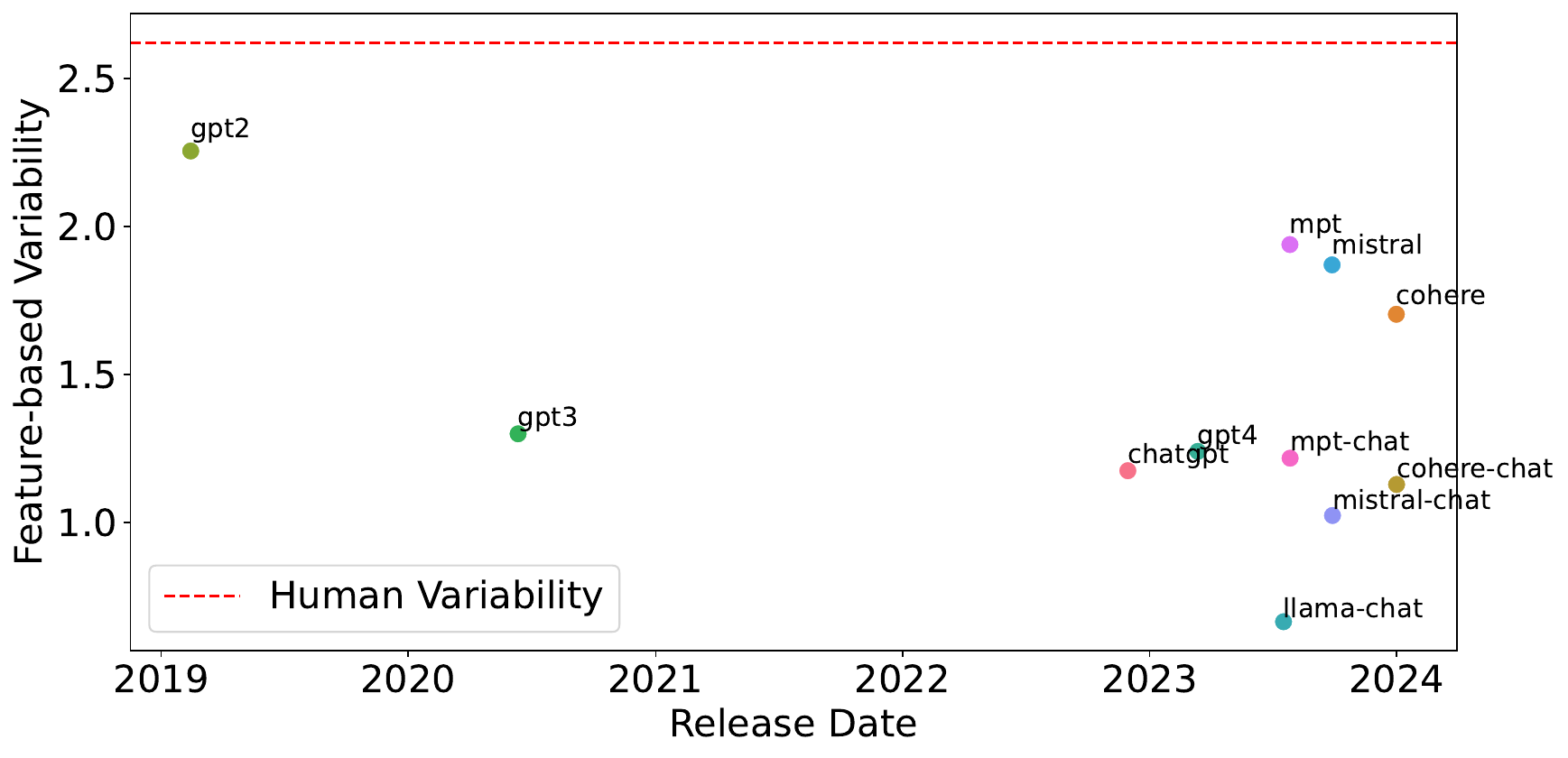}
    \caption{Model Variability with linguistic features in the entire dataset across different release years}
    \label{fig:model_variability_vs_release_date}
\end{figure}

\begin{figure}[ht]
    \centering
    \includegraphics[width=\linewidth]{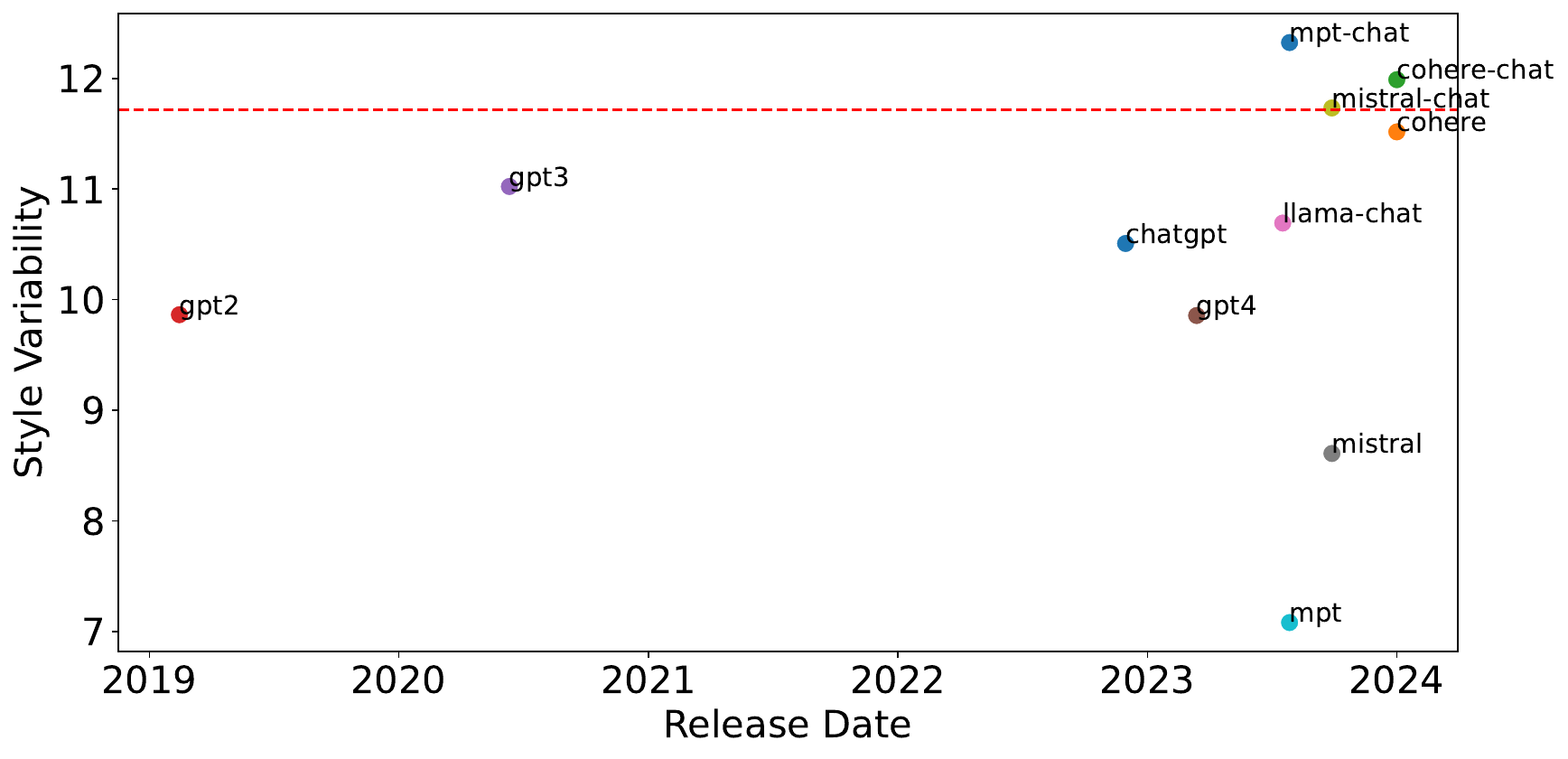}
    \caption{Model Variability with style embeddings in the entire dataset across different release years}
    \label{fig:style_model_variability}
\end{figure}

\subsection{Homogenization of Model Outputs}

Figure \ref{fig:model_variability_vs_release_date} presents the variability of linguistic features across different models and their release dates. The results indicate a trend where MGT exhibit lower linguistic variability compared to HWT and score increasingly similarly to one another, suggesting a homogenization of linguistic styles of models for our set of features. Despite differences in the overall magnitude of variability, Figure \ref{fig:style_model_variability} shows that machine-generated texts exhibit similar variability in style embeddings both among themselves and in comparison to human-written texts. However, the chat models tend to exhibit variability comparable to HWT and higher than their non-chat versions, as expected given the different design of the models.

\section{Discussion} \label{p:dis}
Our analysis of linguistic features highlights different tendencies between human-written (HWT) and machine-generated texts (MGT) in the RAID dataset. Our feature importance analysis shows that human-written texts tend to have less complex syntactic structures and more varied lexical and semantic content. The tendency to produce less complex syntactical structures is in line with previous studies \citep{munoz2024contrasting}. Our variability analysis shows that models tend to produce outputs that are very similar between themselves and are less varied than humans. We attribute this result to the importance of individual style in human-written texts. Nevertheless, models that differ in decoding strategies and repetition penalties produce outputs that reflect their intended design. 
Moreover, the domain variability analysis suggests an influence of linguistic constraints in less flexible and less creative domains such as abstracts \citep{biber2019register}. Notably, newer models tend to be similar in terms of variability between themselves for our linguistic features, possibly pointing to the phenomenon of ``model collapse'' \citep{shumailov2023curse}, where models tend to exhibit low variance due to training on generated data. Analyses based on style embeddings reveal differing levels of variability between chat models and non-chat models, with the former exhibiting variability comparable to that of humans, while the latter show lower variability. These results can be attributed to the effect of reinforcement learning from human feedback, which is used to train chat models.

\section{Conclusion and Future Work}
In our analysis of linguistic features of human-written (HWT) and machine-generated texts (MGT), we show different linguistic tendencies of human-written texts and how recent models tend to generate texts with linguistic characteristics similar to human-written texts, but exhibit a lower variability within themselves based on our set of features. 
Future work should explore diverse corpora with varying characteristics to verify these differences across different domains and languages. Expanding the range of linguistic features, especially those related to content such as the use of metaphors or figurative language, could provide deeper insights. Moreover, our setting of Human/Machine authorship attribution should be expanded from a binary human/non-human setting to a multi-class classification, where detection models have to attribute authorship to humans and to different LLMs.

\section{Limitations}

One limitation of our study lies in its generalizability. We rely on a dataset covering eight domains in the English language. Thus, the applicability of our findings is limited when considering broader linguistic variations across domains and languages. The eight domains covered in our dataset provide valuable insights, but they may not be representative of all possible linguistic contexts characteristic of all textual domains. Furthermore, texts in various languages may have unique linguistic features that limit the relevance of our results to non-English contexts.

While this study identifies and measures a range of linguistic features relevant for comparing human- and machine-generated texts, it does not examine how interactions between these features might provide additional insights. This represents a notable limitation, as individual features may not only operate independently but also in combination. However, analyzing such interactions poses methodological challenges due to the large number of possible feature combinations, especially in the absence of prior hypotheses or a theoretical framework indicating which interactions are likely to be salient.

Our analysis focuses on a set of LLMs that may already have been superseded by more advanced versions due to the rapid advancements in the field. This poses a challenge for the temporal validity of our findings as future LLMs could exhibit different linguistic patterns.

The RAID dataset does not include metadata identifying different human authors of the sampled texts. As a result, we are unable to analyze potential stylistic variations among human writers.

Extending this study to multiple languages, datasets, and model versions could potentially enhance the applicability of our findings across broader contexts and evolving technologies. However, such an extension would require significant data curation efforts and depend on the availability of multilingual linguistic pipelines capable of characterizing texts at scale. Nevertheless, this study and our findings can serve as a foundation for further exploration of LLM outputs.

\section{Ethical Considerations}
In developing our approach, we acknowledge the potential for unintended bias, particularly against non-native English speakers. Some of the linguistic features we analyze may capture characteristics in texts produced by English language learners. This overlap raises important ethical questions.
It is crucial to emphasize that our primary objective is to advance the theoretical understanding of language patterns in texts generated by humans and LLMs, rather than to create tools for real-world applications. The features and techniques described in this paper are intended for research purposes and should not be directly applied in practical systems without careful consideration of their broader implications. Any potential real-world application would require extensive additional research and safeguards. We strongly caution against using these features or similar approaches in high-stakes decision-making processes or in any context where they could disadvantage individuals based on their language competency.

\section*{Acknowledgments}
S.E.Z. is supported by the Deutsche Forschungsgemeinschaft (DFG -- German Research Foundation) under Germany's Excellence Strategy -- EXC-2035/1 -- 390681379. S.A. acknowledges the support of ERC Advanced Grant 101020961 PRODEMINFO.

\bibliography{custom}

\appendix

\section{Classifier results}
This section presents the results from the logistic classifier and the support vector machine (SVM) trained on a set of linguistic features. We also provide a qualitative error analysis for the logistic classifier, illustrating examples of texts that were misclassified.

\subsection{Linguistic features Classifier}
We use a set of linguistic features we calculated to build a logistic classifier for assessing feature importance in distinguishing HWT and MGT. Table \ref{tab:l_classification_performance} provides the details of the overall results.
The classifier achieves an accuracy of 0.61 across all domains. While we do not aim to compare our results against specific baselines or prior studies, it is worth noting that our accuracy scores are consistent with findings from related work on human/machine authorship attribution using classifiers based solely on linguistic features, which typically report performance in the 0.5–0.6 range \citep{alecakir-etal-2024-groningen, sharma-mansuri-2024-team}.

\begin{table}[htbp]
    \centering
    \begin{tabular}{lcc}
        \toprule
        \textbf{Metric} & \textbf{Non-Human} & \textbf{Human} \\
        \midrule
        Accuracy & \multicolumn{2}{c}{0.61} \\
        \midrule
        Precision & 0.62 & 0.60 \\
        Recall    & 0.56 & 0.65 \\
        F1-score  & 0.59 & 0.62 \\
        \bottomrule
    \end{tabular}
    \caption{Classification performance of the logistic regression classifier based on linguistic features}
    \label{tab:l_classification_performance}
\end{table}

\begin{figure}[htbp]
    \centering
    \includegraphics[width=\linewidth]{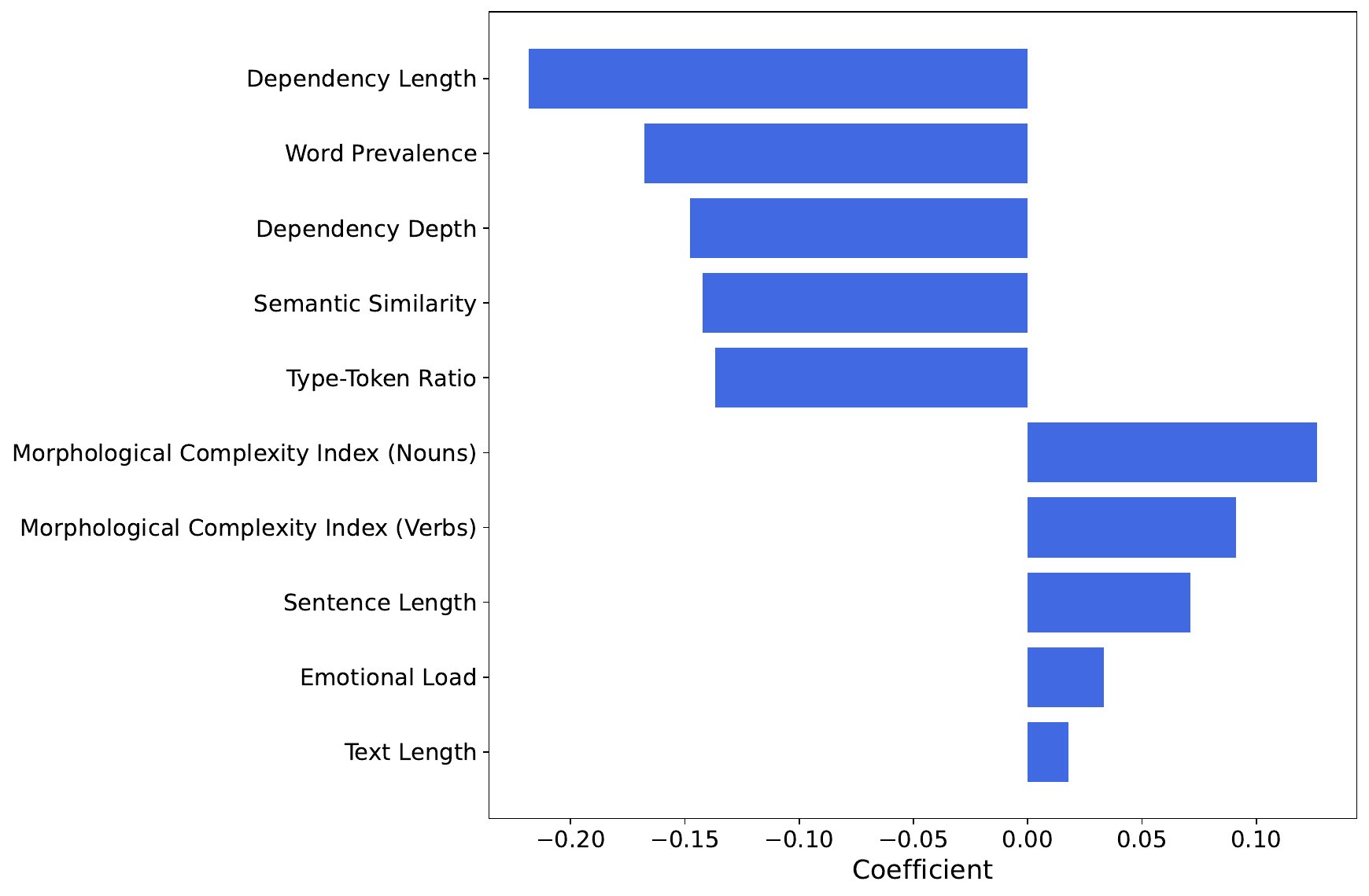}
    \caption{Feature importance for the SVM classifier across all domains. A positive coefficient means that as the feature value increases, the model is more likely to predict human authorship, while a negative coefficient indicates that higher values make the model less likely to predict human authorship.}
    \label{fig:svm_feature_importance}
\end{figure}

In order to demonstrate the robustness of these results, we build a second classifier using an SVM on our set of linguistic features. Table \ref{tab:svm_classification_performance} shows the results from the SVM classifier in distinguishing HWT and MGT. The classifier achieves an accuracy of 0.60 across all domains, in line with the results from the logistic classifier. Moreover, Figure \ref{fig:svm_feature_importance} shows the results from the feature importance analysis of the SVM model. The pattern of the feature importance analysis for the SVM classifier is very similar to that of the logistic classifier (see Figure \ref{fig:feature_importance} in Section \ref{results}). Particularly, all features maintain the same direction,  while the coefficients are slightly lower for the SVM classifier.

\begin{table}[htbp]
    \centering
    \begin{tabular}{lcc}
        \toprule
        \textbf{Metric} & \textbf{Non-Human} & \textbf{Human} \\
        \midrule
        Accuracy & \multicolumn{2}{c}{0.60} \\
        \midrule
        Precision & 0.62 & 0.59 \\
        Recall    & 0.55 & 0.66 \\
        F1-score  & 0.58 & 0.62 \\
        \bottomrule
    \end{tabular}
    \caption{ Classification performance of the Support Vector Machine classifier using linguistic features}
    \label{tab:svm_classification_performance}
\end{table}

\subsection{Qualitative Error Analysis}

\begin{figure}[htbp]
    \centering
    \includegraphics[width=\linewidth]{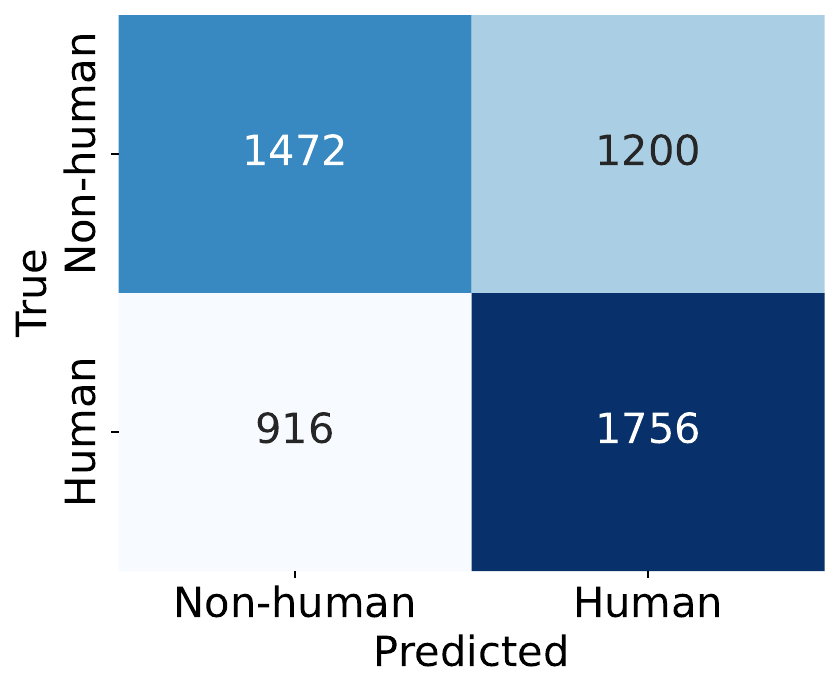}
    \caption{Confusion matrix showing classification performance of the logistic regresson classifier for human vs. machine-generated texts}
    \label{fig:confusion_matrix}
\end{figure}

We present a qualitative error analysis of the logistic classifier by examining misclassified texts. Figure~\ref{fig:confusion_matrix} displays the confusion matrix, while Table~\ref{tab:exqualer} provides three examples of false positives (machine-generated texts predicted as human-written texts) and false negatives (HWT predicted as MGT). To explore whether some specific features systematically drive these errors, we use Cohen’s $d$ \citep{cohen2013statistical} to measure the standardized mean difference between misclassified and correctly classified cases. We focus on false positives (FP) versus true negatives (TN) and false negatives (FN) versus true positives (TP). This allows us to identify the features that likely drive misclassification.

Table \ref{tab:fpvstn} shows that false positives are primarily associated with texts that are longer and exhibit greater morphological complexity (both verbs and nouns) compared to true negatives. Thus, when machine-generated texts are long and structurally rich, the classifier tends to mistake them for HWT.

In contrast, Table \ref{tab:fnvstp} shows that false negatives often involve human-written texts that are shorter and morphologically simpler than true positives. Additionally, misclassified HWT tend to show higher semantic similarity, as human-written texts tend to show greater semantic diversity than MGT, according to our feature importance analysis of the logistic regression classifier.

Future work could investigate whether such cases are also challenging for human annotators, potentially leading to similar misclassification.

\begin{table}[htbp]
\centering
\begin{tabular}{lrrr}
\toprule
\textbf{Feature} & \textbf{Cohen's $d$} \\
\midrule
Text Length & 1.20 \\
Morph. Complexity (Verbs) & 0.88\\
Morph. Complexity (Nouns) & 0.81  \\
Semantic Similarity & -0.66  \\
Emotional Load & 0.52 \\
Type--Token Ratio & -0.45  \\
Dependency Depth & -0.22 \\
Sentence Length & -0.20  \\
Dependency Length & -0.18 \\
Word Prevalence & -0.16 \\
\bottomrule
\end{tabular}
\caption{Cohen's $d$ for linguistic features comparing false positives and true negatives.
Positive values indicate features that are more salient in machine-generated texts (MGT) that were misclassified as human-written (HWT).}
\label{tab:fpvstn}
\end{table}

\begin{table}[htbp]
\centering
\begin{tabular}{lrrr}
\toprule
\textbf{Feature} & \textbf{Cohen's $d$} \\
\midrule
Morph. Complexity (Verbs) & -1.07  \\
Semantic Similarity & 0.92  \\
Morph. Complexity (Nouns) & -0.75  \\
Emotional Load & -0.61 \\
Text Length & -0.58 \\
Type--Token Ratio & 0.57 \\
Dependency Length & 0.46  \\
Dependency Depth & 0.44  \\
Sentence Length & 0.28  \\
Word Prevalence & 0.15 \\
\bottomrule
\end{tabular}
\caption{Cohen's $d$ for linguistic features comparing false negatives and true positives. Positive values indicate features that are more salient in HWT misclassified as MGT.
}
\label{tab:fnvstp}
\end{table}

\begin{table*}[htbp]
\centering
\begin{tabularx}{\textwidth}{X}
\toprule
\multicolumn{1}{c}{\textbf{False Positives (machine-generated predicted as human-written texts)}} \\
\midrule

Hey fellow Redditors,

I'm on day 35 of my CT (camel trail) and I'm starting to feel a little stuck. I've been following the same routine for weeks now, and I can't help but wonder if I'm clinging to my nightmare because it's all I know at this point. [...]. \\ 
\hline
In this paper, we explore the connection between Brunet-Derrida particle systems, free boundary problems, and Wiener-Hopf equations. We first introduce the concept of Brunet-Derrida particle systems, which are stochastic systems that describe the evolution of a collection of particles that interact with each other through a non-local competition mechanism. [...]. \\
\hline
The trees are all so naked, and shivering with cold! They clatter over one another like drunken men when there's wind; and the snow lies hard on their arms, as if it were trying to get them into bed—trying too hard, for they don’t move an inch in reply….. [...]. \\

\midrule
\multicolumn{1}{c}{\textbf{False Negatives (human-written predicted as machine-generated texts)}} \\
\midrule

David Innes and his captive, a member of the reptilian Mahar master race of the interior world of Pellucidar, return from the surface world in the Iron Mole invented by his friend and companion in adventure Abner Perry. Emerging in Pellucidar at an unknown location, David frees his captive. [...]. \\
\hline
The Children of Zion, published in January 1998, is considered as a documentary that was based on a collection of fragments of records compiled in Palestine in 1943 by the Eastern Center for Information, a Polish government group. [...]. \\
\hline
Newspaper columnist Mitch Albom recounts time spent with his 78-year-old sociology professor, Morrie Schwartz, at Brandeis University, who was dying from Lou Gehrig's disease (ALS). Albom, a former student of Schwartz, had not corresponded with him since attending his college classes 16 years earlier. [...]. \\

\bottomrule
\end{tabularx}
\caption{Examples of misclassified texts by the logistic regression classifier} \label{tab:exqualer}
\end{table*}

\clearpage

\section{Prompt Examples} \label{PE}
In this section, we present examples of prompts from the RAID dataset used to generate MGT.  
All models rely on the same specific set of prompts. For instance:

\begin{itemize}
    \item \texttt{Write the abstract for the academic paper titled ``Model Theory for a Compact Cardinal.''}
    \item \texttt{Write a recipe for ``Anise Toasts Recipe.''}
\end{itemize}

\section{Positive and Negative Emotions}
\begin{figure}[htbp]
    \centering
    \includegraphics[width=\linewidth]{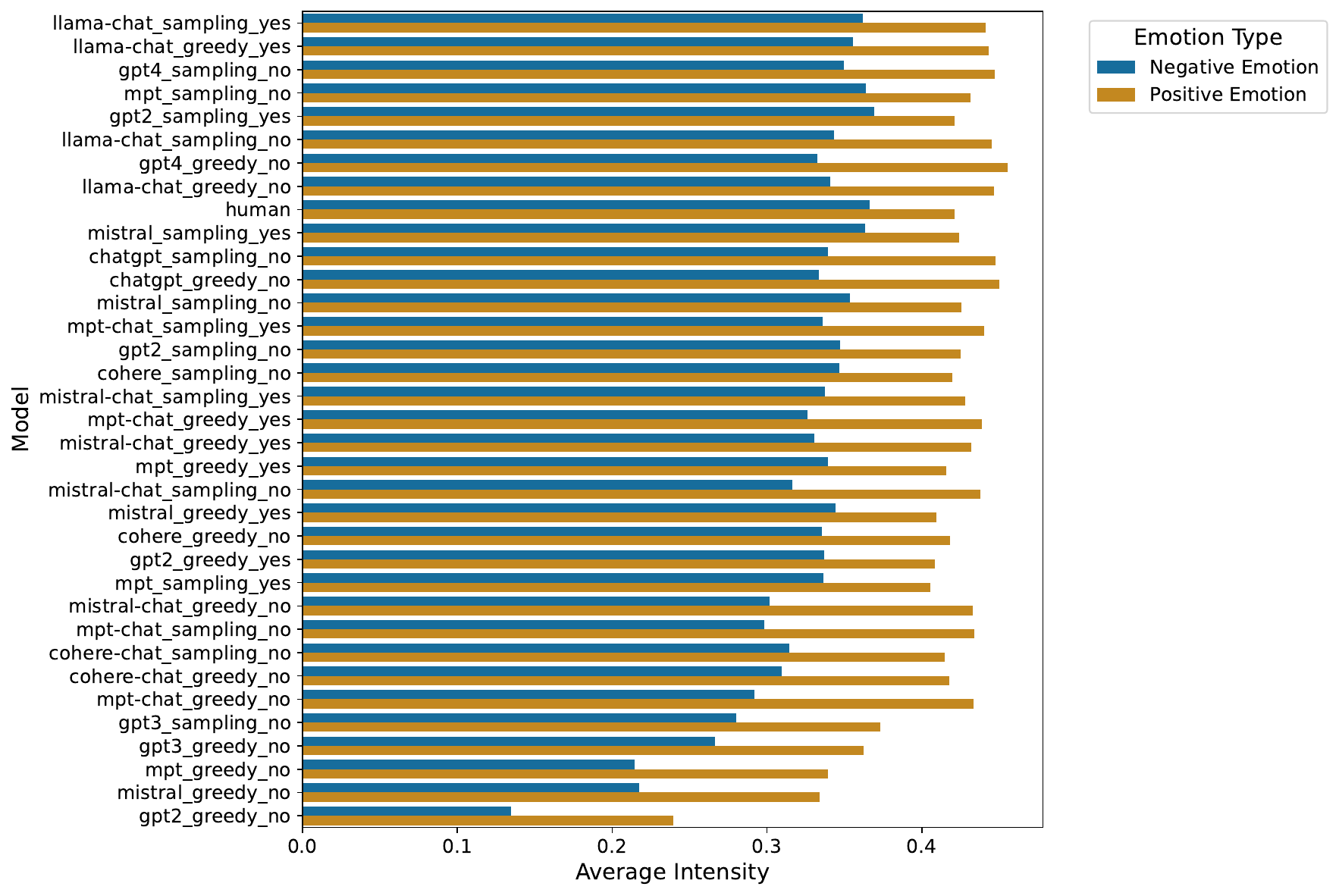}
    \caption{Negative and Positive Emotion Intensity per Model. This figure presents the emotional intensity across humans and different models.}
    \label{fig:raid_emotion_intensity}
\end{figure}

In Figure \ref{fig:raid_emotion_intensity}, we show the differences in positive and negative emotion intensities of different models with their decoding strategy and the presence or not of a penalty on repetition. Indeed, HWT score high on negative emotion intensity, while more recent models tend to use fewer negative emotion words, arguably due to their alignment with human and/or machine feedback.

\section{Domain variability with style embeddings}

\begin{figure}[ht]
    \centering
    \includegraphics[width=\linewidth]{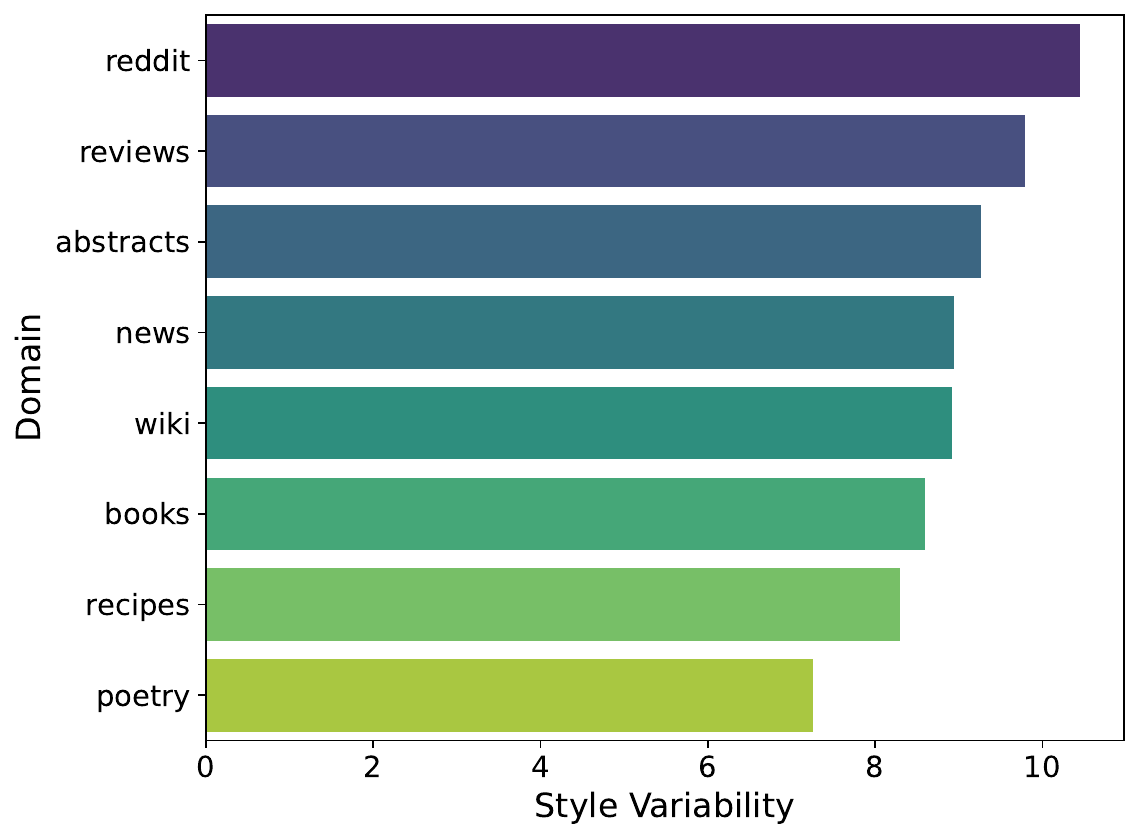}
    \caption{Domain variability with style embeddings in the entire dataset}
    \label{fig:style_domain_variability}
\end{figure}

\begin{figure}[ht]
    \centering
    \includegraphics[width=\linewidth]{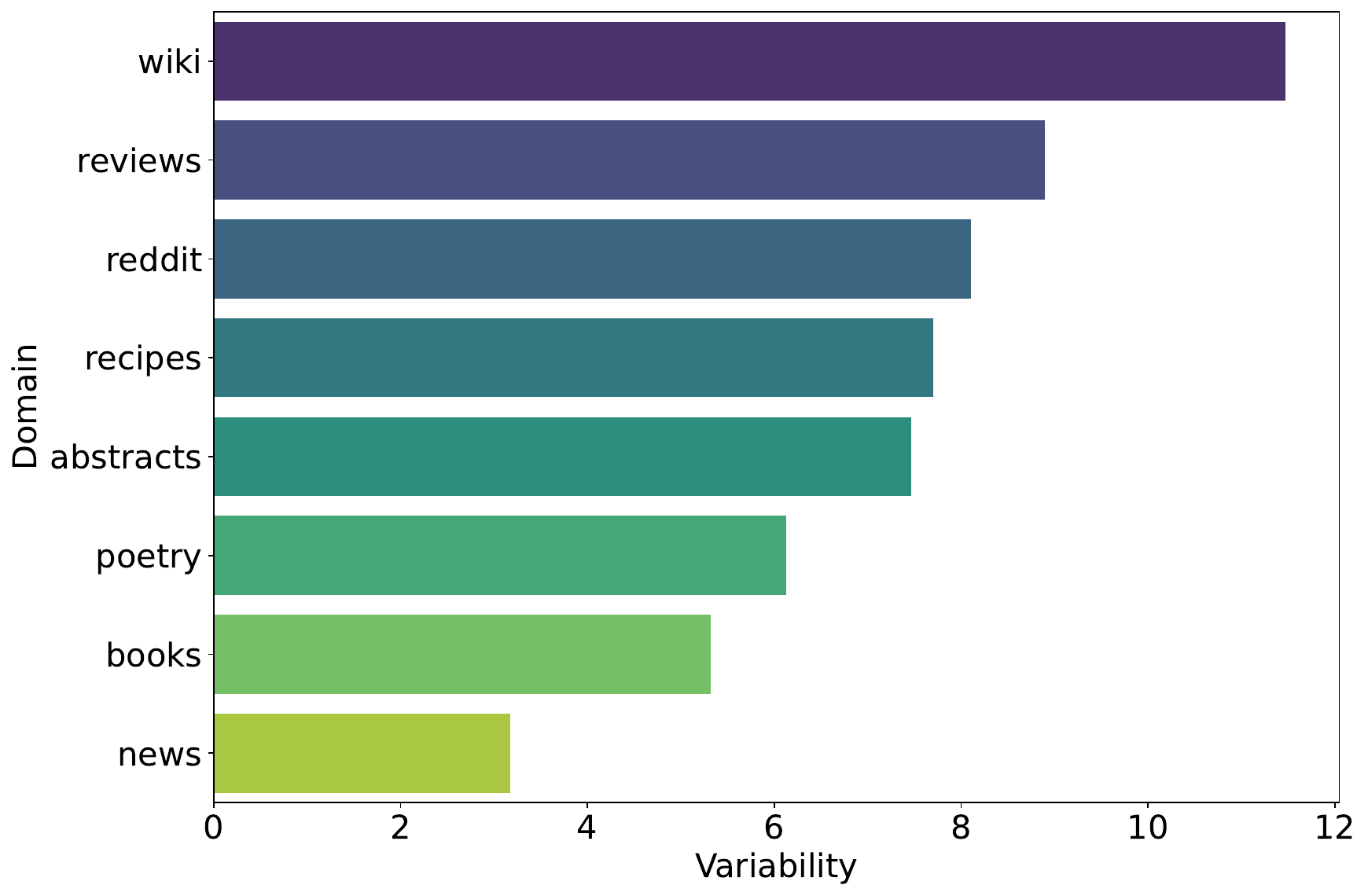}
    \caption{Domain variability with style embeddings in the entire dataset for humans}
    \label{fig:style_domain_variability_huamns}
\end{figure}

\begin{figure}[ht]
    \centering
    \includegraphics[width=\linewidth]{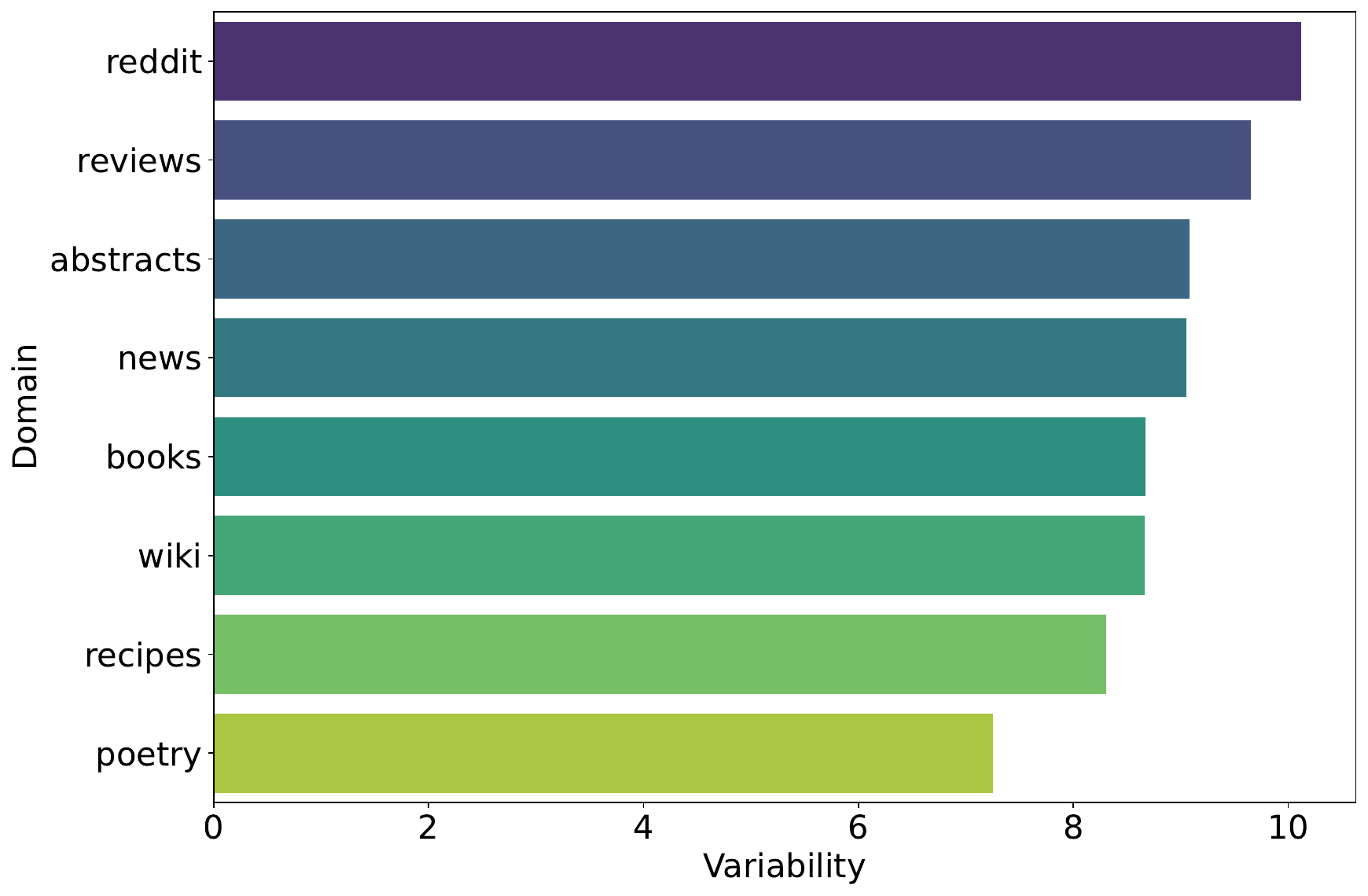}
    \caption{Domain variability with style embeddings in the entire dataset for non-humans}
    \label{fig:style_domain_variability_nonhumans}
\end{figure}

Figure \ref{fig:style_domain_variability} shows the variability within domains for HWT and MGT based on style embeddings for the entire datasets. Moreover, Figure \ref{fig:style_domain_variability_huamns} and Figure \ref{fig:style_domain_variability_nonhumans} disentangle the variability respectively of HWT and MGT. We argue that differences in variability between linguistic features and style embeddings stem from the different representations that are considered for the analyses. Future research should explore potential explanations for the differing variability rankings across text domains.

\section{Linguistic Features Visualization} \label{LFV}
\begin{figure*}[htbp]
    \centering
    \includegraphics[width=\textwidth]{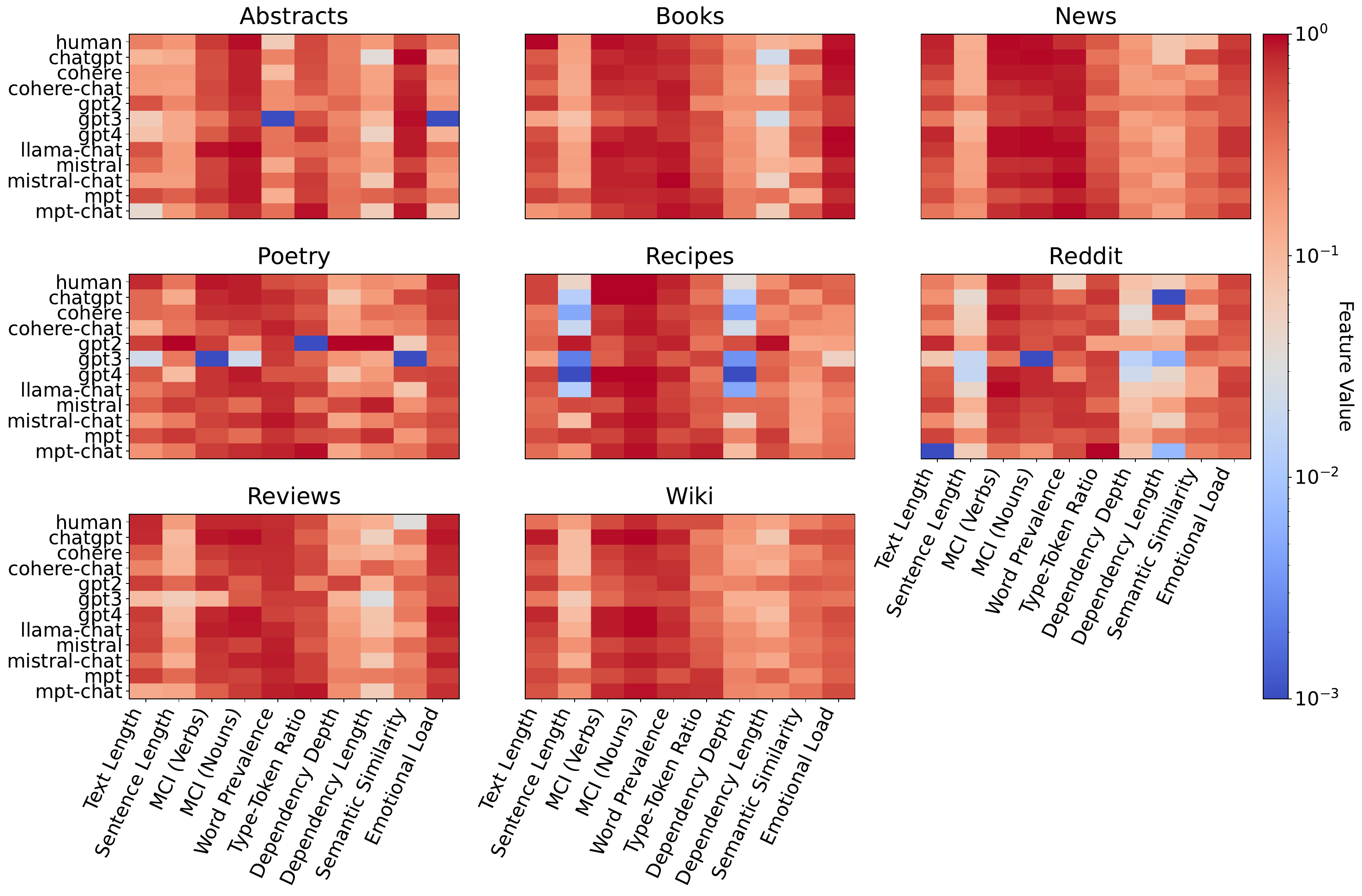}
    \caption{Multi-panel heatmap displaying log-normalized feature values across different text domains and models}
    \label{fig:multi_panel_heatmap_mean}
\end{figure*}

Figure \ref{fig:multi_panel_heatmap_mean} shows the mean values of the linguistic features (standard deviation in Figure \ref{fig:multi_panel_heatmap_std}) across domains for each model in the dataset. The following linguistic features are considered: Text Length, Sentence Length, Morphological Complexity Index (MCI) for nouns, Morphological Complexity Index (MCI) for verbs, dependency depth, dependency length, word prevalence, type-token ratio, semantic similarity and emotionality. 
Notably, we observe a difference in variability between domains, especially when dealing with more ``creative'' domains such as Poetry and Books (See Figures \ref{fig:multi_panel_heatmap_mean} and \ref{fig:multi_panel_heatmap_std}). Especially, human-written texts show greater differences in average text length within these domains.

\paragraph{Poetry Domain}
In the poetry domain, on aggregate, HWT score highest in average text length and emotionality, while maintaining fairly simpler syntactic structures than most models. We can also notice that human-written texts tend to score higher in morphological complexity for verbs and nouns.

\paragraph{Books Domain}
In the books domain, human-written texts score highest in average text length. However, HWT score lower on emotionality than some models, and they employ more complex syntactic structures than in Poetry. Nevertheless, we notice that HWT tend to score high in morphological complexity for verbs and nouns in this domain as well.

\paragraph{Abstracts Domain}
For the abstracts domain, the patterns are different in comparison with poetry and books domains. Human-written texts score lower on average text length and emotionality than the other two domains. We can notice how recent models (e.g., GPT4) behave extremely more consistently and more similar to HWT than their older models (e.g., GPT2) in text length, sentence length, syntactic depth and length, and also in semantic similarity.

\paragraph{Recipes Domain}
In our set of linguistic features, the recipes domain shows that human-written texts score lowest in dependency length, underlying the tendency to employ simpler syntactic structures than MGT. This tendency involves semantic similarity as well, where HWT appear extremely more consistent in the semantic content than in previously discussed domains.

\paragraph{Reddit Domain}
In the Reddit domain, HWT score lower on morphological complexity than in previously discussed domains. This could point to differences in language register (e.g., less formal vs more formal) between a user-based domain with no genre-specific constraints and more strict domains \citep{biber2019register}. Again, newer models show more similar patterns to human-written texts than their older versions, especially in semantic similarity (e.g., GPT).

\paragraph{Reviews Domain}
The reviews domain shows that human-written texts employ a slightly higher use of frequently known words than in the other discussed domains, as we can see in Figure \ref{fig:multi_panel_heatmap_mean}.

\paragraph{Wikipedia Domain}
In the Wikipedia domain, HWT score very low in average text length, as well as in morphological complexity and emotionality, in comparison with the other discussed domains. Clearly, we can observe the effect of very rigid genre-specific constraints and how humans have to strictly abide by them. 

\paragraph{News Domain}
The news domain shows that HWT score high in average text length, morphological complexity, and fairly high in emotionality in comparison to most models. 
Unlike in other domains, human-written texts also score lowest in word prevalence, indicating the use of less common words, and lower in semantic similarity, suggesting a more varied semantic content.

\section{Experimental Setup Details} \label{apx:setupDetails}

All experiments were carried out with an NVIDIA A100 GPU (40GB memory) and standard CPU resources. We did not train or fine-tune any models; instead, we relied on pre-trained encoders purely for inference:

\begin{itemize}
    \item \textbf{StyleDistance}, based on RoBERTa-base with approximately 125M parameters.
    \item \textbf{paraphrase-MiniLM-L6-v2}, a MiniLM encoder with approximately 22M parameters.
\end{itemize}

The calculation of linguistic features (e.g., syntactic depth, sentence length, morphological complexity, emotion intensity) was performed using \texttt{spaCy} (\texttt{en\_core\_web\_sm}) and other lexicon-based methods on CPU. Only the transformer-based components (style embeddings and semantic similarity) were executed on the GPU, requiring about 3 GPU hours.
The overall computational budget was therefore modest, with CPU-bound feature extraction accounting for most of the runtime.

\end{document}